\documentclass[10pt,twocolumn,letterpaper]{article}

\usepackage[pagenumbers]{cvpr} %

\usepackage[pagebackref,breaklinks,colorlinks,allcolors=cvprblue]{hyperref}
\usepackage{graphicx}
\usepackage{booktabs}
\usepackage{pifont}
\usepackage{multicol}
\usepackage{multirow}
\usepackage{colortbl}  %
\usepackage{amssymb}   %
\usepackage{fontawesome5} %
\usepackage{makecell}
\usepackage{listings}
\usepackage{floatpag}
\floatpagestyle{empty}

\definecolor{cvprblue}{rgb}{0.21,0.49,0.74}
\definecolor{mgreen}{RGB}{57, 181, 74}
\newcommand{\tablestyle}[2]{\setlength{\tabcolsep}{#1}\renewcommand{\arraystretch}{#2}\centering\footnotesize}

\newcommand{\cmark}{\textcolor[HTML]{228B22}{\ding{51}}}
\newcommand{\xmark}{\textcolor[HTML]{DC3545}{\ding{55}}}

\newcommand{\best}[1]{\cellcolor[rgb]{0.957,0.71,0.706}#1}
\newcommand{\besttwo}[1]{\cellcolor[rgb]{0.976,0.859,0.718}#1}
\newcommand{\bestthree}[1]{\cellcolor[rgb]{1,1,0.613}#1}
\newcommand{\bestfour}[1]{\cellcolor[rgb]{1,1, 0.853}#1}

\usepackage{amsmath,amsfonts,bm}

\def\eqref#1{equation~\ref{#1}}

\def\1{\bm{1}}

\def\rmI{{\mathbf{I}}}

\def\vpi{{\bm{\pi}}}

\DeclareMathAlphabet{\mathsfit}{\encodingdefault}{\sfdefault}{m}{sl}
\SetMathAlphabet{\mathsfit}{bold}{\encodingdefault}{\sfdefault}{bx}{n}

\def\gI{{\mathcal{I}}}

\def\gV{{\mathcal{V}}}

\def\paperID{22} %
\def\confName{ICCV}
\def\confYear{2025}
\def\OURS{\textsc{Seva}}
\def\ours{\OURS\xspace}

\title{
\textsc{Stable Virtual Camera}: 
Generative View Synthesis with Diffusion Models
}

\author{
Jensen (Jinghao) Zhou$^{1,2,*,\dagger}$~~~~
Hang Gao$^{1,3,*,\dagger}$\\%~~~~
Vikram Voleti$^1$~~~~
Aaryaman Vasishta$^1$~~~~
Chun-Han Yao$^1$~~~~
Mark Boss$^1$\\%~~~~
Philip Torr$^2$~~~~
Christian Rupprecht$^2$~~~~
Varun Jampani$^1$\\
\vspace{-6pt}\\
$^1$Stability AI~~~~
$^2$University of Oxford~~~~
$^3$University of California, Berkeley\\
\vspace{-6pt}\\
\href{https://stable-virtual-camera.github.io}{{\includegraphics[height=0.4cm]{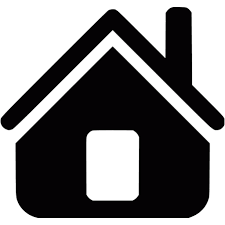}}}~~~~
\href{https://stability.ai/news/introducing-stable-virtual-camera-multi-view-video-generation-with-3d-camera-control}{{\includegraphics[height=0.4cm,width=0.4cm]{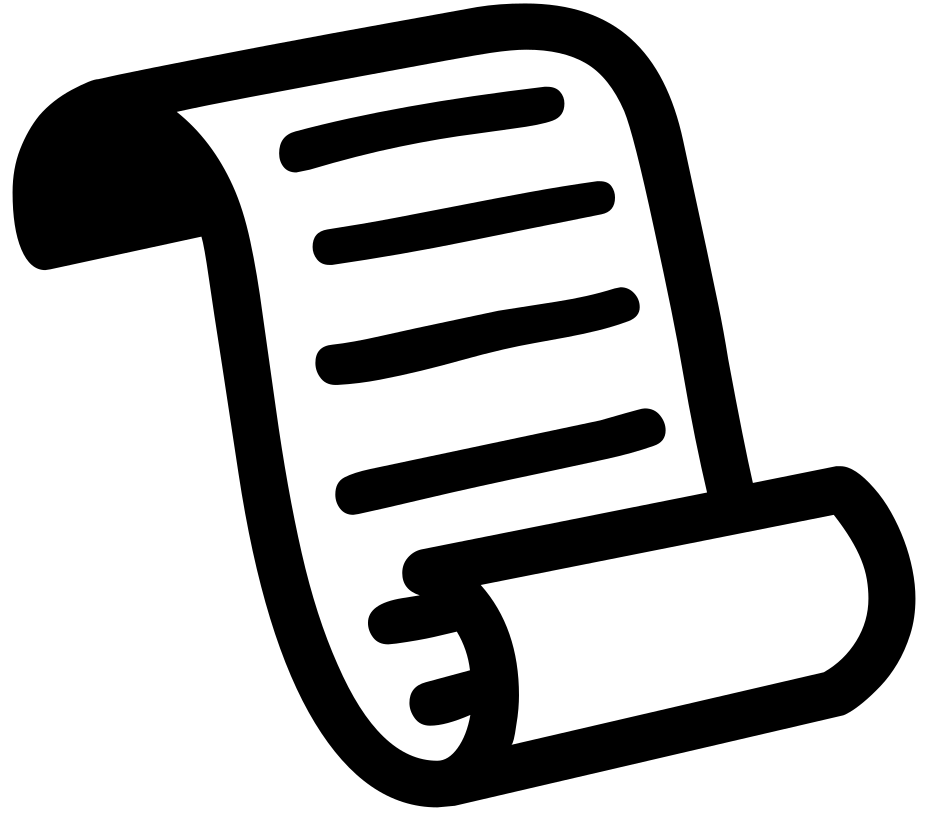}}}~~~~
\href{https://github.com/Stability-AI/stable-virtual-camera}{{\includegraphics[height=0.4cm]{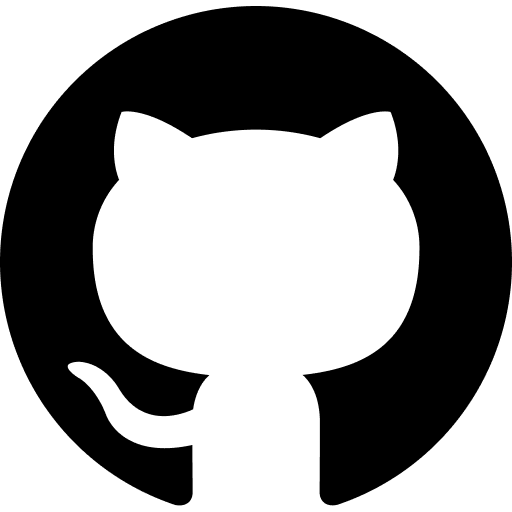}}}~~~~
\href{https://huggingface.co/stabilityai/stable-virtual-camera}{{\includegraphics[height=0.4cm]{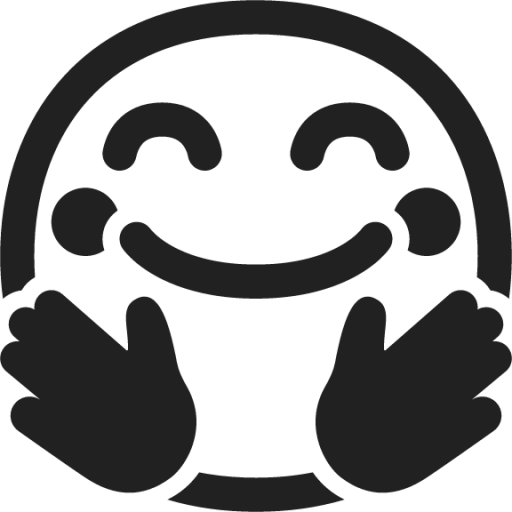}}}~~~~
\href{https://huggingface.co/spaces/stabilityai/stable-virtual-camera}{{\includegraphics[height=0.4cm]{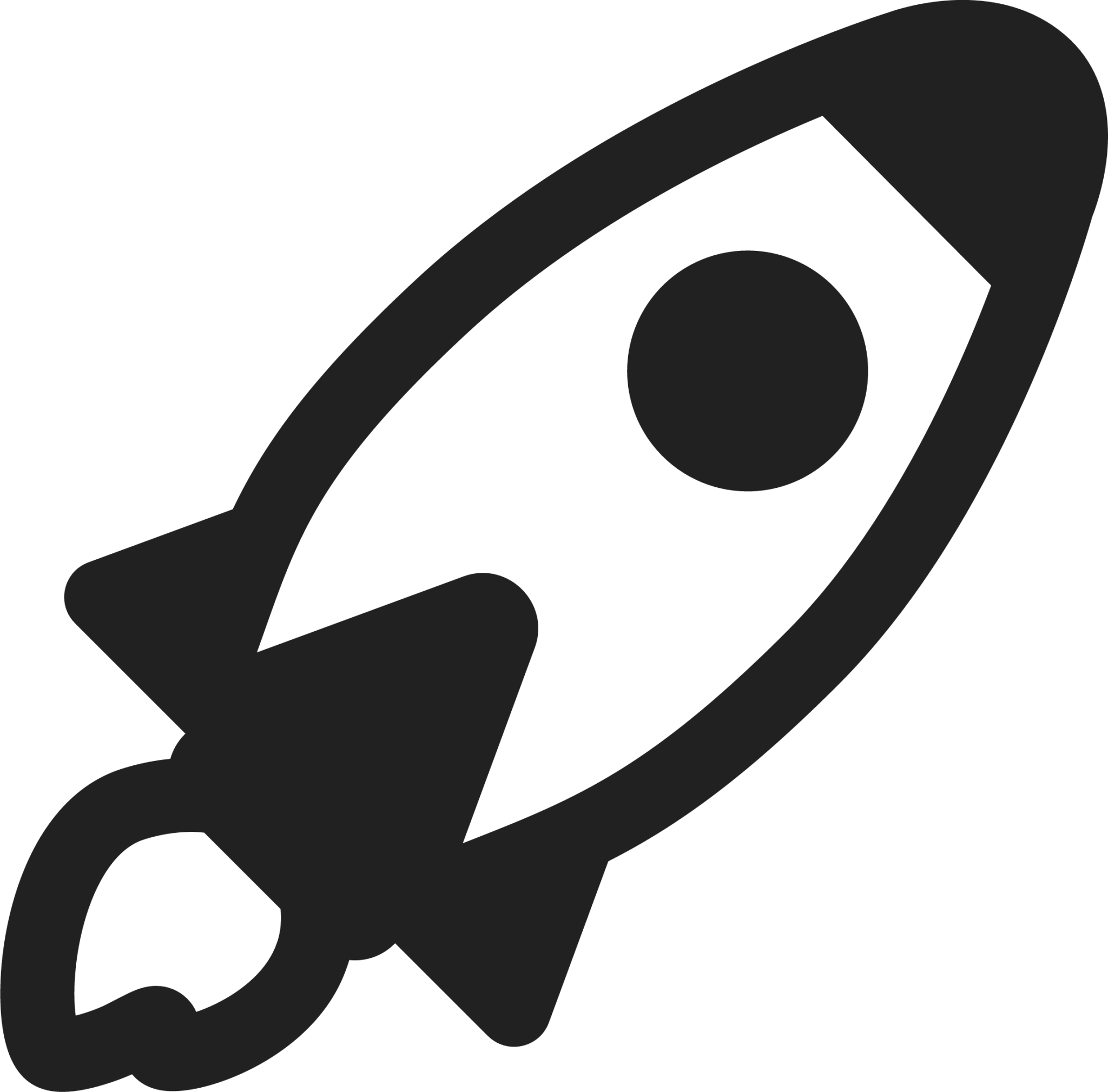}}}~~~~
\href{https://www.youtube.com/channel/UCLLlVDcS7nNenT_zzO3OPxQ}{{\includegraphics[height=0.4cm]{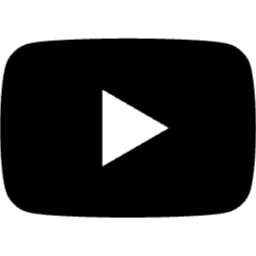}}}
}

\begin{document}
\twocolumn[
{
\renewcommand\twocolumn[1][]{#1}%
\maketitle
\begin{center}
    \vspace{-0.4cm}
    \includegraphics[width=\textwidth]{fig/assets/teaser.pdf}
    \captionof{figure}{
        \textbf{Generative view synthesis.}
        \textsc{Stable Virtual Camera} generates novel views from any number of input views and target cameras, which the user can specify anywhere.
        We show three examples: single view with simple orbit camera trajectory (top); two views with long camera trajectory (middle); and nine views with large spatial range (bottom).
        Please visit our website for video samples.
    }
    \label{fig:teaser}
\end{center}
}]

\renewcommand\thefootnote{} %
\footnotetext{$^*$Equal contribution.}
\footnotetext{$^\dagger$Jensen and Hang did the work during internships at Stability AI.}
\renewcommand\thefootnote{\arabic{footnote}} %

\begin{abstract}
We present \textsc{\underline{S}tabl\underline{e} \underline{V}irtual C\underline{a}mera} (\ours), a generalist diffusion model that creates novel views of a scene, given any number of input views and target cameras.
Existing works struggle to generate either large viewpoint changes or temporally smooth samples, while relying on specific task configurations.
Our approach overcomes these limitations through simple model design, optimized training recipe, and flexible sampling strategy that generalize across view synthesis tasks at test time.
As a result, our samples maintain high consistency without requiring additional 3D representation-based distillation, thus streamlining view synthesis in the wild.
Furthermore, we show that our method can generate high-quality videos lasting up to half a minute with seamless loop closure.
Extensive benchmarking demonstrates that \ours outperforms existing methods across different datasets and settings.
\end{abstract}
  
\section{Introduction}
\label{sec:intro}
Novel view synthesis (NVS) aims to generate realistic, 3D-consistent images of a scene from arbitrary camera viewpoints given any number of camera-posed input views. Traditional methods, which rely on dense input views, treat NVS as a 3D reconstruction and rendering problem~\cite{mildenhall2020nerf,mueller2022instant,kerbl3Dgaussians}, but this approach fails with sparse inputs. 
Generative view synthesis addresses this limitation by leveraging modern deep network priors~\cite{yu2020pixelnerf,poole2022dreamfusion}, enabling immersive 3D interactions in uncontrolled environments without the need to capture large image sets per scene.
In this work, we focus on generative view synthesis and, unless otherwise specified, refer to it simply as NVS for clarity.

Despite recent progress~\cite{liu2023zero,reconfusion,cat3d,yu2024viewcrafter,wang2024motionctrl,jin2024lvsmlargeviewsynthesis,4dim}, NVS in the wild remains limited due to two key challenges:
First, existing methods struggle to generate both large viewpoint changes~\cite{wang2024motionctrl,yu2024viewcrafter} and temporally smooth samples~\cite{liu2023zero,sargent2023zeronvs,reconfusion,cat3d} while being constrained by rigid task configurations, such as a fixed number of input and target views~\cite{reconfusion,jin2024lvsmlargeviewsynthesis,yu2024viewcrafter,4dim}, reviewed in~\cref{tab:review}.
Second, their sampling consistency is often insufficient, necessitating additional NeRF distillation to fuse inconsistent results into a coherent representation~\cite{sargent2023zeronvs,reconfusion,cat3d}.
These limitations hinder their applicability across diverse NVS tasks, which we address in this work.

\begin{figure*}[p]
    \includegraphics[width=\textwidth]{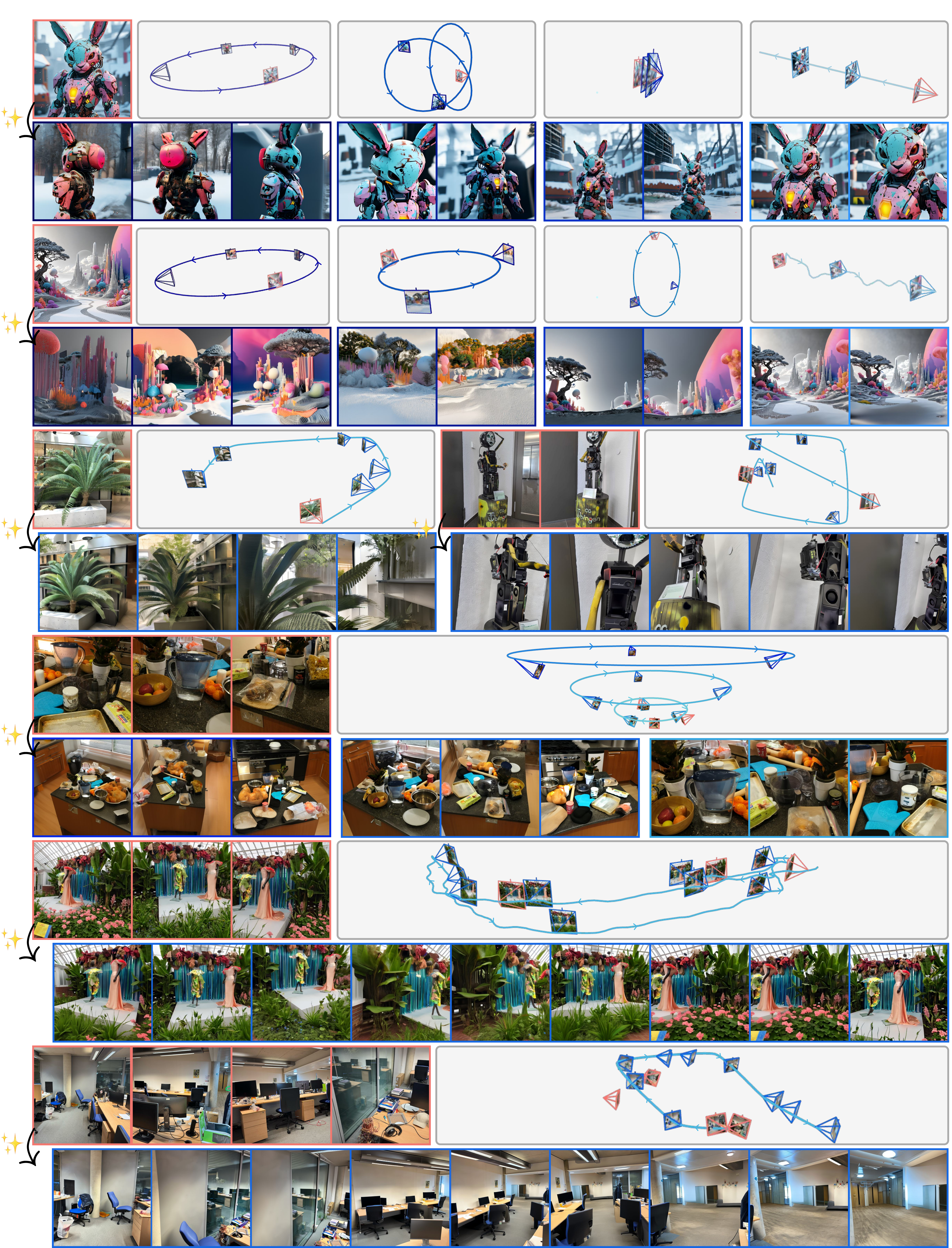}
    \vspace{-0.5cm}
    \caption{
        \textbf{Diverse camera control.}
        \ours generates photorealistic novel views following diverse camera trajectories.
        This includes orbit, spiral, zoom out, dolly zooms, and any user-specified trajectories.
        Please visit our website for more visual results.
    }
    \label{fig:fancy}
    \vspace{-0.5cm}
\end{figure*}

We present \textsc{\underline{S}tabl\underline{e} \underline{V}irtual C\underline{a}mera}\footnote{
We name this model in tribute to the Virtual Camera~\cite{virtualcamerawiki} cinematography technology, a pre-visualization technique to simulate real-world camera movements.
} (\ours), a diffusion-based NVS model that generalizes across a spectrum of view synthesis tasks without requiring NeRF distillation.
With a single network, \ours generates high-quality novel views that strike both large viewpoint changes and temporal smoothness, while supporting any number of input and target views. 
Our approach simplifies the NVS pipeline without requiring distillation from a 3D representation, thus streamlining it for real-world applications. 
For the first time, we demonstrate high-quality videos lasting up to half a minute with precise camera control and seamless loop closure in 3D.
We highlight these results in~\cref{fig:teaser} and showcase more examples of camera control in~\cref{fig:fancy}.

To achieve this, we carefully design our pipeline in three key aspects: model design, training recipe, and sampling method at inference.
First, \ours avoids explicit 3D representations within the network, allowing the model to inherit strong priors from pre-trained 2D models.
Second, during training, we carefully craft our view selection strategy to cover both small and large viewpoint changes, ensuring strong generalization to diverse NVS tasks.
Third, at inference, we introduce a two-pass procedural sampling approach that supports flexible input-target configurations. 
Together, these design choices create a versatile 3D ``virtual camera simulation system'' capable of synthesizing novel views along arbitrary camera trajectories with any number of input and target views, without using a 3D representation.

We conducted a unified benchmark across 10 datasets and a variety of experimental settings, including both open-source and proprietary models. 
Our benchmark reflects the diversity of real-world NVS tasks across the board and systematically evaluates existing methods beyond their comfort zones. 
We find that \ours consistently outperforms previous works, achieving +1.5 dB PSNR over the state of the art CAT3D~\cite{cat3d} in its own setup. 
Moreover, our method generalizes well to in-the-wild user captures, with input views ranging from 1 to 32. 

In summary, our key contributions with the 
\ours model include:
(1) a training strategy for jointly modeling large viewpoint changes and temporal smoothness,
(2) a two-pass procedural sampling method for smooth video generation along arbitrary long camera trajectories,
(3) a comprehensive benchmark that evaluates NVS methods across different datasets and settings, and
(4) an open-source release of model weights to support future research.

\begin{table}
    \tablestyle{0.8pt}{1.0}
    \centering
    \begin{tabular}{lcccc}
        \toprule
        \multirow{2}{*}{model} & training & generation & interpolation & input \\
        & data & capacity & smoothness & flexibility \\
        \midrule
        \multicolumn{5}{l}{\textbf{Regression-based}} \\
        pixelNeRF~\cite{yu2020pixelnerf} & \faCube & \xmark & \cmark & sparse (1) \\
        pixelSplat~\cite{charatan23pixelsplat} & \faMountain & \xmark & \cmark & sparse (2) \\
        MVSplat~\cite{chen2025mvsplat} & \faMountain & \xmark & \cmark & sparse (2) \\
        Long-LRM~\cite{longlrm} & \faMountain & \xmark & \cmark & semi-dense (\{16, 32\}) \\
        LVSM~\cite{jin2024lvsmlargeviewsynthesis} & \faCube \ / \faMountain & \xmark & \cmark & sparse (\{2, 4\}) \\
        \midrule
        \multicolumn{5}{l}{\textbf{Diffusion-based: image models}}\\
        Zero123~\cite{liu2023zero} & \faCube & \cmark & \xmark & sparse (1) \\
        ZeroNVS~\cite{sargent2023zeronvs} & \faCube \ \faMountain & \cmark & \xmark & sparse (1) \\
        ReconFusion~\cite{reconfusion} & \faCube \ \faMountain & \cmark & \xmark & sparse (3) \\
        CAT3D~\cite{cat3d} & \faCube \ \faMountain & \cmark & \xmark & sparse ([1, 9]) \\
        \midrule
        \multicolumn{5}{l}{\textbf{Diffusion-based: video models}} \\
        SV3D~\cite{voleti2024sv3d} & \faCube & \xmark & \cmark & sparse (1) \\
        MotionCtrl~\cite{wang2024motionctrl} & \faMountain & \xmark & \cmark & sparse (1) \\
        ViewCrafter~\cite{yu2024viewcrafter} & \faMountain & \xmark & \cmark & sparse (2) \\
        4DiM~\cite{4dim} & \faMountain & \cmark & \cmark & sparse (\{1, 2, 8\}) \\
        \midrule
        \ours & \faCube \ \faMountain & \cmark & \cmark & \makecell{ sparse ([1, 8]), \\ semi-dense ([9, 32$^*$]) } \\
        \bottomrule
    \end{tabular}
    \vspace{-0.1cm}
    \caption{
    \textbf{Comparison of existing NVS models} based on the source of training data and key attributes.
    \ours is trained on both object-level (\faCube) and scene-level (\faMountain) data, offering flexible input conditioning, strong generation capacity, and smooth view interpolation.
    We define generation capacity and interpolation smoothness of each work based on their evaluation setting and our benchmark results.
    $^*$This upper-bound can be up to hundreds for dense captures, we test our model up to 32 views in practice.
    }
    \label{tab:review}
    \vspace{-0.5cm}
\end{table}

\thispagestyle{empty}

\section{Background}
\label{sec:background}
We consider the evaluation of an NVS model across three key criteria:
(1) generation capacity---the ability to synthesize missing regions for large viewpoint changes; 
(2) interpolation smoothness---the ability to produce seamless transitions between views;
and
(3) input flexibility---the ability to handle a variable number of input and target views; 
We review existing NVS models based on these criteria in~\cref{tab:review}, including the types of training data.

\subsection{Types of NVS Tasks}
\label{sec:nvstasks}
\begin{figure}
    \includegraphics[width=\linewidth]{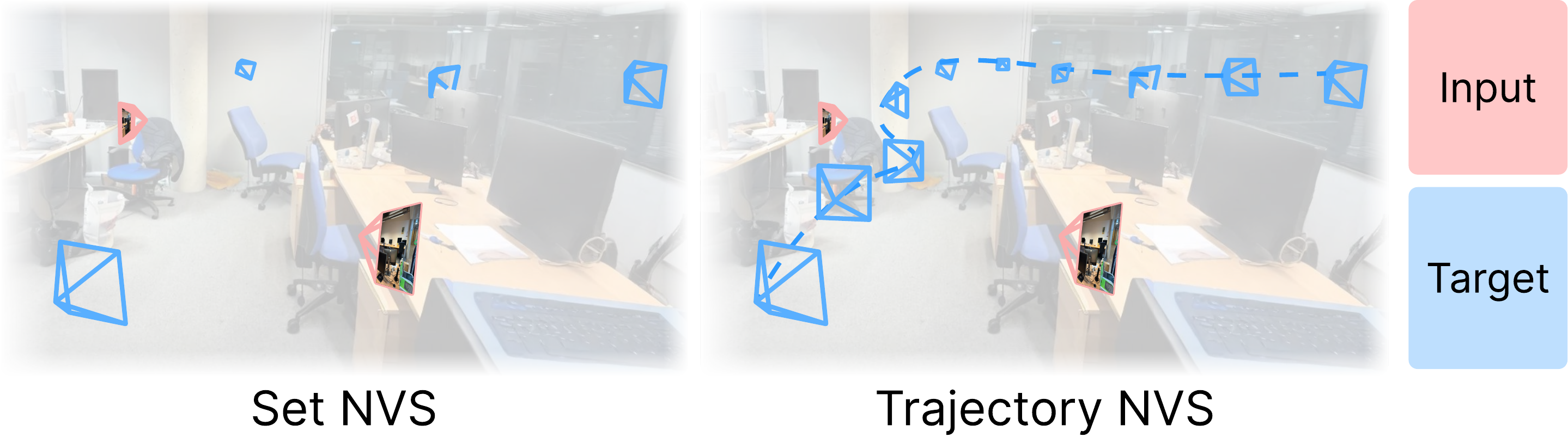}
    \vspace{-0.6cm}
    \caption{  
        \textbf{Set NVS \textit{versus} trajectory NVS.} Set NVS generates target views as an image set, whereas trajectory NVS produces them as a trajectory video.
    }
    \vspace{-0.3cm}
    \label{fig:set_trajectory}
    \vspace{-0.3cm}
\end{figure}

Given $M$ input view images $\rmI^\text{inp} \in \mathbb{R}^{M\times H \times W \times 3}$ of $H \times W$ resolution, along with their corresponding cameras $\vpi^\text{inp}$, NVS involves predicting $N$ targets views $\rmI^\text{tgt} \in \mathbb{R}^{N\times H \times W \times 3}$, specified by their respective cameras $\vpi^\text{tgt}$.
For each camera, we assume we know both intrinsics and extrinsics.
Based on the number of input views, we define the ``sparse-view regime'' as having up to 8 input views, and the ``semi-dense-view regime'' as an intermediate state bridging the sparse-view regime and dense captures, which typically involve hundreds of views.
Based on the nature of their target views, we bucket a broad range of NVS tasks into ``set NVS'' and ``trajectory NVS'', as shown in~\cref{fig:set_trajectory}.

\vspace{-0.3cm}
\paragraph{Set NVS} considers a set of target views in arbitrary order, usually across a large spatial range. The order of views is often not helpful here, and a good NVS model requires great generation capacity to excel at this task.
We note that some works address only this task (\eg ReconFusion~\cite{reconfusion}).

\vspace{-0.3cm}
\paragraph{Trajectory NVS} regards target views along a smooth camera trajectory, such that they form a video sequence.
However, they are often sampled within a small spatial range in a shorter video.
To solve this task, a good NVS model requires great interpolation smoothness to produce consistent and non-flickering results. 
We note that some existing works address only this task (\eg ViewCrafter~\cite{yu2024viewcrafter}).

\subsection{Existing Models}
\label{sec:nvsmodels}

We group existing approaches into regression- and diffusion-based models based on their high-level design choices.
A more detailed discussion of related works can be found in~\cref{sec:related}.

\vspace{-0.3cm}
\paragraph{Regression-based models} learn a deterministic mapping:
\begin{equation*}
    f_\theta\Big(\rmI^\text{inp}, \vpi^\text{inp}, \vpi^\text{tgt}\Big),
\end{equation*}
to directly generate $\rmI^\text{tgt}$ deterministically from $\rmI^\text{inp}, \vpi^\text{inp}, \vpi^\text{tgt}$. 
$f_\theta$ can be either an end-to-end network parameterized by $\theta$, or 
a composition of a feed-forward prediction of an intermediate 3D representation and then a neural renderer~(\eg, NeRF~\cite{mildenhall2021nerf} or 3DGS~\cite{kerbl3Dgaussians}). 
For the latter case, set NVS and trajectory NVS are solved in the same way since there exists a persistent 3D representation.

\vspace{-0.3cm}
\paragraph{Diffusion-based models} capture the conditional distribution:
\begin{equation*}
    p_\theta\Big(\rmI^\text{tgt}\ |\ \rmI^\text{inp}, \vpi^\text{inp}, \vpi^\text{tgt}\Big), 
\end{equation*}
from which $\rmI^\text{tgt}$ are sampled~\cite{ho2020denoising} iteratively.
We highlight two types of models within this scope: Image and Video models.
Image models are trained on unordered image sets, such that $(\rmI^\text{inp}, \rmI^\text{tgt}) \sim \gI$, where $\gI = \{\rmI_{\sigma(1)}, \rmI_{\sigma(2)}, \cdots, \rmI_{\sigma(M+N)}\}$ is an image batch, and $\sigma(\cdot)$ is a random permutation function, where camera parameters are omitted for simplicity. 
Image models usually thrive at set NVS, but struggle in trajectory NVS since they are designed to generate images and not videos.
Additionally, the unordered nature of all views solicits flexible input conditioning.
Video models are instead trained on ordered views, such that $(\rmI^\text{inp}, \rmI^\text{tgt}) \sim \gV$, where $\gV = \{\rmI_1, \rmI_2, \cdots, \rmI_{M+N}\}$ is a randomly sampled video batch with ordering preserved. 
Additional temporal operators may also be used to improve the temporal smoothness, such as temporal positional encoding and temporal attention.
In contrast with image models, video models thrive at trajectory NVS, but struggle in set NVS.
Moreover, all existing video models require both input and target views to be ordered (input views followed by target ones), constraining their input flexibility~\cite{svd,voleti2024sv3d,wang2024motionctrl,he2024cameractrl,he2022lvdm}.

\subsection{Remarks and Motivation}
\looseness=-1
Existing tasks pose critical challenges to our design choices.
Specifically, our design choices are made to achieve high generation capacity, smooth view interpolation, and flexible input conditioning, as compared in~\cref{tab:review}.
In this way, we can employ a single model for both tasks, described next.

\section{Method}
\label{sec:method}
\begin{figure*}
    \includegraphics[width=\linewidth]{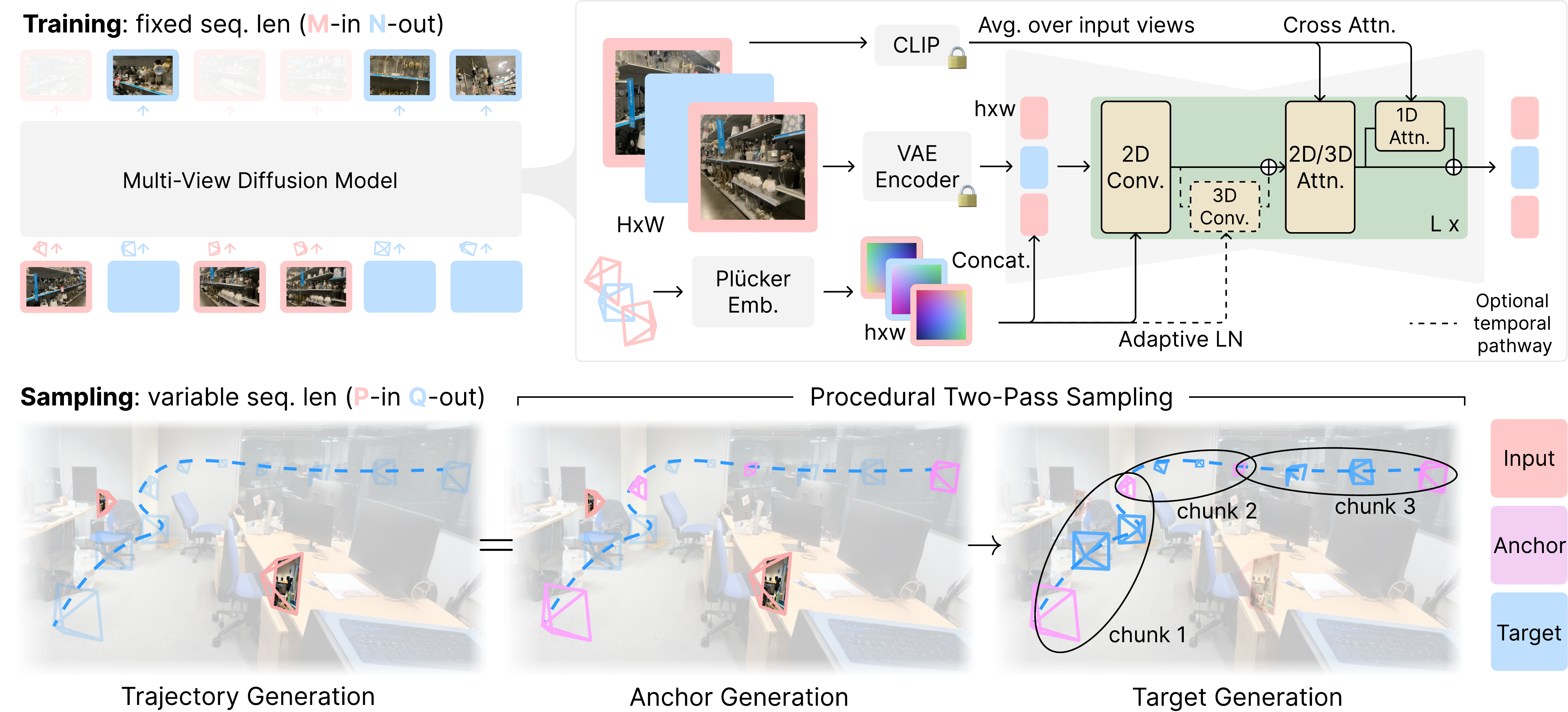}
    \vspace{-0.5cm}
    \captionof{figure}{
        \textbf{Method.}
        \ours is trained with fixed sequence length as a ``$M$-in $N$-out'' multi-view diffusion model with standard architecture. It conditions on CLIP embeddings, VAE latents of the input views, and their corresponding camera poses.
        During sampling, \ours can be cast as a generative ``$P$-in $Q$-out'' renderer that works with variable sequence length, where $P$ and $Q$ need not be equal to $M$ and $N$. To enhance temporal and 3D consistency across generated views, especially when generating along a trajectory, we present procedural two-pass sampling as a general strategy.
    }
    \label{fig:system}
    \vspace{-0.5cm}
\end{figure*}

We describe our model design and training strategy in \cref{sec:model}, then our sampling process at test time in \cref{sec:sampling,sec:sampling_special}.
A system overview is provided in \cref{fig:system}.

\subsection{Model Design and Training}
\label{sec:model}

We consider a ``$M$-in $N$-out'' multi-view diffusion model $p_\theta$, as notated in~\cref{sec:nvsmodels}. 
We formulate this learning problem as a standard diffusion process~\cite{ho2020denoising} without any change.

\vspace{-0.3cm}
\paragraph{Architecture.}
\looseness=-1
Our model is based on the publicly available SD 2.1~\cite{stablediffusion}, which consists of an auto-encoder and a latent denoising U-Net.
Following~\cite{cat3d}, we inflate the 2D self-attention of each low-resolution residual block into 3D self-attention~\cite{shi2023mvdream} within the U-Net.
To improve model capacity, we add 1D self-attention along the view axis after each self-attention block via skip connection~\cite{resnet,largesgd}, bumping the model parameters from 870M to 1.3B.
Optionally, we further tame this model into a video model by introducing 3D convolutions after each residual block via skip connection, similar to~\cite{blattmann2023align,svd}, yielding 1.5B total parameters. The temporal pathway can be enabled during inference when frames within one forward pass are known to be spatially ordered, enhancing output's smoothness.

\vspace{-0.3cm}
\paragraph{Conditioning.}
To fine-tune our base model into a multi-view diffusion model, we add camera conditioning as Pl\"ucker embedding~\cite{plucker1865xvii} via concatenation~\cite{cat3d} and adaptive layer normalization~\cite{zheng2023free3d}.
We normalize $\vpi^\text{inp}$ and $\vpi^\text{tgt}$ by first computing the relative pose with respect to the first input camera and then normalizing the scene scale such that all camera positions are within a $[-2, 2]^3$ cube.
For each input frame, we first encode its latent then concatenate with its Pl\"ucker embedding and a binary mask~\cite{cat3d,svd} differentiating between input and target views. For each target frame, we use the noisy state of its latent instead.
Additionally, we find it helpful~\cite{voleti2024sv3d} to also inject high-level semantic information via CLIP~\cite{radford2021clip} image embedding.
We zero initialize new weights for additional channels in the first layer.
In our experiment, we found that our model can quickly adapt to these conditioning changes and produce realistic images with as few as $5K$ iterations.

\vspace{-0.3cm}
\paragraph{Training.}
Let us define the training context window length $T = |\rmI^\text{inp}| + |\rmI^\text{tgt}| = M + N$.
One natural goal is to support large $T$ such that we can generate a larger set of frames.
However, we find that naive training is prone to divergence, and we thus employ a two-stage training curriculum.
During the first stage, we train our model with $T = 8$ with a batch size of $1472$ for $100K$ iterations.
In the second stage, we train our model with $T = 21$ with a batch size of $512$ for $600K$ iterations.
Given a training video sequence, we randomly sample the number of input frames $M \in [1, T-1]$ and the frames $(\rmI^\text{inp}, \rmI^\text{tgt})$. We find it important to jointly sample $\rmI$ with a smaller subsampling stride to ensure sufficient temporal granularity and avoid missing critical transitions with a small probability ($0.2$ is used in practice). In the optional video training stage, we only train temporal weights with data sampled with a small subsampling stride and a batch size of $512$ for $200K$ iterations.
We shift the signal-to-noise ratio (SNR) in all stages as more noise is necessary to destroy the information when training with more frames, corroborating findings from~\cite{hoogeboom2023simple,esser2024scaling,cat3d}.
The model is trained with squared images with $H=W=576$.

\subsection{Sampling Novel Views}
\label{sec:sampling}
\looseness=-1
Once the diffusion model is trained, we can sample it for a wide range of NVS tasks during test time.
Formally, let us consider a ``$P$-in $Q$-out'' NVS task during testing, where we are given~$P=|\rmI^\text{inp}|$ input frames and aim to produce~$Q=|\rmI^\text{tgt}|$ target frames. 
Our goal is to design a generic sampling strategy that works for all~$P$ and~$Q$ configurations, where $P$ and $Q$ need not be equal to $M$ and $N$.

We make two key observations:
First, within a single forward pass, predictions are 3D consistent,  provided the model is well-trained.
Second, when $P + Q > T$, $\rmI^\text{tgt}$ must be split into smaller chunks of size $Q_i$ such that $P + Q_i \leqslant T$ for the $i^\text{th}$ forward pass. We term this practice \textit{one-pass sampling}.
However, predictions across these forward passes would be inconsistent unless they share common frames to maintain local consistency within a spatial neighborhood.
Building on these observations, we summarize our sampling process under two scenarios: $P + Q \leqslant T$ and $P + Q > T$.

\vspace{-0.3cm}
\paragraph{\boldsymbol{$P + Q \leqslant T$}.}
We fit the task within one forward pass for simplicity and consistency.
As shown in~\cref{app:addexperiments}, we find it works better to pad the forward pass to have exactly $T$ frames by repeating the first input image, compared to changing the context window $T$ zero-shot.

\vspace{-0.3cm}
\paragraph{\boldsymbol{$P + Q > T$}.}
We propose \textit{procedural two-pass sampling}:
In the first pass, we generate anchor frames $\rmI^\text{acr}$ using all input frames $\rmI^\text{inp}$.
In the second pass, we divide $\rmI^\text{tgt}$ into chunks and generate them using $\rmI^\text{acr}$ (and optionally $\rmI^\text{inp}$) according to the spatial distribution of $\rmI^\text{acr}$ and $\rmI^\text{tgt}$.
Given the distinct nature of the two tasks of interest---set NVS and trajectory NVS---\eg, differences in the availability of views' ordering, we design tailored chunking strategies for each task. %

For set NVS, 
we consider $\mathrm{nearest}$ procedural sampling.
We first generate $\rmI^\text{acr}$ based on pre-defined trajectory priors, similar to~\cite{cat3d}, \eg, 360 trajectories for object-centric scenes, or spiral trajectories for forward-facing scenes.
We then divide $\rmI^\text{tgt}$ into chunks \wrt $\rmI^\text{acr}$ using nearest neighbor.
Specifically, the $i^\text{th}$ forward pass involves:
\begin{equation*}
    \mathrm{nearest}: \{\rmI^\text{acr}_i\} \cup \{\rmI^\text{tgt}_j \mid \mathrm{NN}(\rmI^\text{tgt}_j, \rmI^\text{acr}) = \rmI^\text{acr}_i\}.
\end{equation*}
We considered two strategies of procedural sampling: $\mathrm{nearest}$ as described above, and $\mathrm{gt + nearest}$ strategy by appending $\rmI^\text{inp}$ into each forward pass.
We find that the $\mathrm{gt + nearest}$ strategy performs better than $\mathrm{nearest}$ and thus default to it instead.
In the absence of trajectory priors, we revert to one-pass sampling. In practice, employing nearest anchors enhances qualitative consistency, albeit on a limited scale.

For trajectory NVS,
we consider $\mathrm{interp}$ procedural sampling.
We first generate a subset of target frames as $\rmI^\text{acr}$ by uniformly spanning the target camera path with a stride $\Delta = \lfloor \frac{Q}{T - 2} \rfloor$.
We then generate the rest of $\rmI^\text{tgt}$ as segments between those anchors:
\begin{equation*}
    \mathrm{interp}: \{\rmI^\text{acr}_i, \rmI^\text{tgt}_{i\cdot\Delta + 1}, \cdots, \rmI^\text{tgt}_{(i + 1)\cdot\Delta - 1}, \rmI^\text{acr}_{i+1}\}.
\end{equation*}
Since the input to the model is ordered, we can leverage temporal weights to further improve smoothness (\cref{sec:exp_trajnvs}). 
Similarly, $\mathrm{gt + interp}$ is possible by appending $\rmI^\text{inp}$ with $\Delta = \lfloor \frac{Q}{T - P - 2} \rfloor$. We find that $\mathrm{interp}$ is sufficiently robust, and choose it as the default option.
The $\mathrm{interp}$ strategy drastically outperforms its counterparts (\eg, one-pass, or $\mathrm{gt + nearest}$ procedural sampling) in terms of temporal smoothness.

\subsection{Scaling Sampling for Large $P$ and $Q$}
\label{sec:sampling_special}

Next, we examine two special cases when $P+Q>T$: $P>T$ and $Q\gg T$. Here, we make a tailored design for anchor generation in the first pass, while keeping target generation in the second pass unchanged.

\vspace{-0.3cm}
\paragraph{\boldsymbol{$P>T$}.} 
In the semi-dense-view regime (\eg,~$P=32$), we extend the context window length $T$ zero-shot to accommodate all $P$ input views and anchor views in one pass during anchor generation.
Empirically, $T$ can even be extended up to hundreds without severe degradation in photorealism in the generated outputs. 
We find that the diffusion model generalizes well in this case as long as the input views cover the majority of the scene, shifting the task from generation to primarily interpolation.
In the sparse-view regime (\ie, $P\leq8$), we observe similar performance degradation caused by zero-shot extension of $T$ compared to what we have found when $P+Q\leqslant T$.
Refer to~\cref{sec:discussions} for a detailed discussion.

\begin{figure}
    \includegraphics[width=\linewidth]{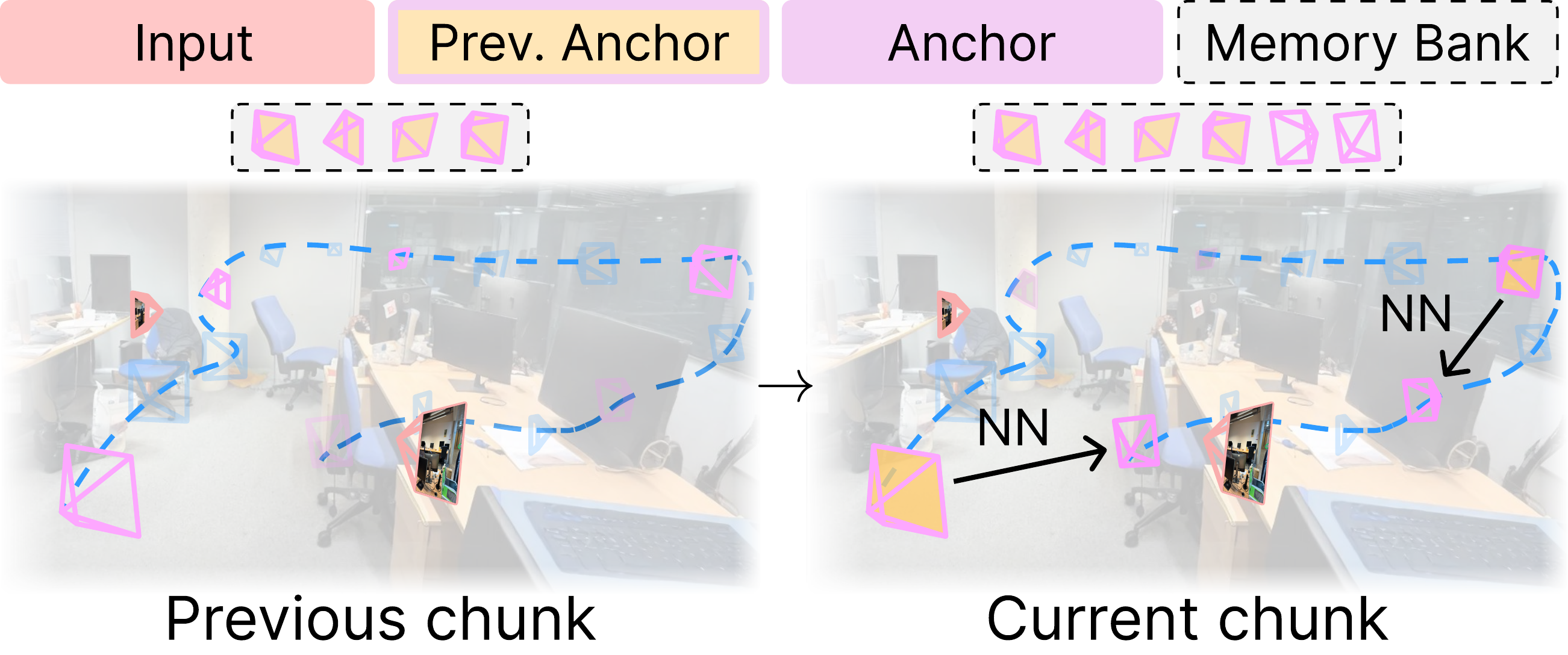}
    \vspace{-0.5cm}
    \captionof{figure}{
        \textbf{Anchor generation when \boldsymbol{$Q>>T$}.} 
        We introduce a memory bank composed of previously-generated anchor views and their corresponding camera poses.
        The lookup of spatial neighbors helps improve long-term 3D consistency.
    }
    \vspace{-0.5cm}
    \label{fig:system_long}
\end{figure}

\vspace{-0.3cm}
\paragraph{\boldsymbol{$Q\gg T$}.}
When the number of target views $Q$ is large, \eg, in large-set NVS or long-trajectory NVS, even anchors will be chunked into different forwards in the first pass, leading to the inconsistency of anchors. 
To this end, we maintain a \textit{memory bank} of anchor views previously generated, as shown in~\cref{fig:system_long}. 
We generate anchors auto-regressively by retrieving their spatially nearest ones from the memory bank, similar to the $\mathrm{nearest}$ strategy introduced above for the second pass. 
In~\cref{sec:exp_trajnvs_long}, we show that this strategy drastically outperforms the standard practice of reusing temporally nearest anchors previously generated in long video literature~\cite{he2022lvdm}, in terms of long-range 3D consistency, especially for hard trajectories.

\section{Experiments}
\label{sec:experiments}
We employ a single model for a spectrum of settings and find that \ours model generalizes well under the three criteria (\cref{tab:review}).
We cover different NVS tasks (set NVS and trajectory NVS) and examine one special task of interest---long trajectory NVS.
We also cover different input regimes (single-view, sparse-view, and semi-dense-view).
A discussion about several key properties is presented in~\cref{sec:discussions}.

\subsection{Benchmark}
\label{sec:benchmark}

\paragraph{Datasets, splits, and the number of input views.}
We consider (1) object datasets, e.g., OmniObject3D~\cite{wu2023omniobject3d} (OO3D) and GSO~\cite{downs2022google}; (2) object-centric scene datasets, e.g., LLFF~\cite{mildenhall2019local}, DTU~\cite{jensen2014large}, CO3D~\cite{co3d}, and WildRGBD~\cite{wildrgbd} (WRGBD); and (3) scene datasets, e.g., RealEstate10K~\cite{realestate10k} (RE10K), Mip-NeRF 360~\cite{barron2021mip} (Mip360), DL3DV140~\cite{dl3dv10k} (DL3DV), and Tanks and Temples~\cite{knapitsch2017tanks} (T\&T).
We consider a wide range of the number of input views $P$, ranging from sparse-view regime to semi-dense-view regime, evaluating models' input flexibility.
To establish a comprehensive and rigorous comparison with baselines, we consider different dataset splits utilized in prior works with the same input-view configuration, unless specified as our split (O).
These include splits used in 4DiM~\cite{4dim} (D), ViewCrafter~\cite{yu2024viewcrafter} (V), pixelSplat~\cite{charatan23pixelsplat} (P), ReconFusion~\cite{reconfusion} (R), SV3D~\cite{voleti2024sv3d} (S), and Long-LRM~\cite{longlrm} (L).
For example, the 4DiM~\cite{4dim} (D) split on the RE10K dataset is 128 out of all 6711 test scenes with $P=1$.

\vspace{-0.3cm}
\paragraph{Small-viewpoint \textit{versus} large-viewpoint NVS.}
Sweeping across all datasets, splits, and input-view configurations reveals a diverse benchmark of setups. To better evaluate models' generation capacity and interpolation smoothness (\cref{sec:nvstasks}), we propose to categorize these setups into two groups---\textit{small-viewpont} NVS and \textit{large-viewpoint} NVS---depending on the disparity between $\rmI^\text{tgt}$ and $\rmI^\text{inp}$. 
small-viewpoint NVS with smaller disparities emphasizes interpolation smoothness and continuity with nearby input views, whereas large-viewpoint NVS with larger disparities requires a model to generate prominent unseen areas from input observations, predominantly assessing models' generation capacity.
See~\cref{tab:benchmark_dataset} for the complete list. Refer to~\cref{app:benchmark} for the detailed choice of datasets, splits, the number of input views, and the way to measure disparity.

\vspace{-0.3cm}
\paragraph{Baselines.}
We consider a range of proprietary models, including ReconFusion~\cite{reconfusion}, CAT3D~\cite{cat3d}, 4DiM~\cite{4dim}, LVSM~\cite{jin2024lvsmlargeviewsynthesis}, and Long-LRM~\cite{longlrm}.
We also consider various open-source models, including SV3D~\cite{voleti2024sv3d}, MVSplat~\cite{chen2025mvsplat}, depthSplat~\cite{depthsplat}, MotionCtrl~\cite{chen2023motion}, and ViewCrafter~\cite{yu2024viewcrafter}. 
As outlined in \cref{sec:nvsmodels}, these baselines encompass both regression-based and diffusion-based approaches, providing a comprehensive framework for comparison.

\subsection{Set NVS}
\label{sec:exp_setnvs}

\begin{table*}
\tablestyle{1.2pt}{1.0}
\centering
\begin{tabular}{lccccccccccccccccccccc}
    \toprule
    \multirow{3}{*}{Method} & dataset & OO3D & GSO & \multicolumn{5}{c}{RE10K} & \multicolumn{2}{c}{LLFF} & \multicolumn{2}{c}{DTU} & \multicolumn{2}{c}{CO3D} & \multicolumn{2}{c}{WRGBD} & Mip360 & \multicolumn{2}{c}{DL3DV} & \multicolumn{2}{c}{T\&T} \\
    \cmidrule(lr){2-2} \cmidrule(lr){3-3} \cmidrule(lr){4-4} \cmidrule(lr){5-9} \cmidrule(lr){10-11} \cmidrule(lr){12-13} \cmidrule(lr){14-15} \cmidrule(lr){16-17} \cmidrule(lr){18-18} \cmidrule(lr){19-20} \cmidrule(lr){21-22}
     & split & O & O & D~\cite{4dim} & V~\cite{yu2024viewcrafter} & P~\cite{charatan23pixelsplat} & \multicolumn{2}{c}{R~\cite{reconfusion}}   & \multicolumn{2}{c}{R~\cite{reconfusion}} & \multicolumn{2}{c}{R~\cite{reconfusion}} & V~\cite{yu2024viewcrafter} & R~\cite{reconfusion} & O$_\mathrm{e}$ & O$_\mathrm{h}$ & R~\cite{reconfusion} & O & L~\cite{longlrm} & V~\cite{yu2024viewcrafter} & L~\cite{longlrm} \\
     \cmidrule(lr){2-2} \cmidrule(lr){3-3} \cmidrule(lr){4-4} \cmidrule(lr){5-5} \cmidrule(lr){6-6} \cmidrule(lr){7-7} \cmidrule(lr){8-9} \cmidrule(lr){10-11} \cmidrule(lr){12-13} \cmidrule(lr){14-14} \cmidrule(lr){15-15} \cmidrule(lr){16-16} \cmidrule(lr){17-17} \cmidrule(lr){18-18} \cmidrule(lr){19-19} \cmidrule(lr){20-20} \cmidrule(lr){21-22}
     & $P$ & 3 & 3 & 1 & 1 & 2 & 1 & 3 & 1 & 3 & 1 & 3 & 1 & 3 & 3 & 6 & 6 & 6 & 32 & 1 & 32 \\
    \midrule
    \multicolumn{20}{l}{\textbf{Regression-based models}} \\
    \multicolumn{2}{l}{Long-LRM~\cite{longlrm}} & - & - & - & - & - & - & - & - & - & - & - & - & - & - & - & - & - & \best{23.86} & - & \best{18.20} \\
     \multicolumn{2}{l}{MVSplat~\cite{chen2025mvsplat}} & \bestthree{\underline{14.78}} & \bestfour{\underline{15.21}} & \bestthree{\underline{20.42}} & \bestthree{\underline{20.32}} & \bestthree{26.39} & \besttwo{\underline{21.56}} & \besttwo{\underline{25.64}} & \bestthree{\underline{11.23}} & \bestfour{\underline{12.50}} & \bestthree{\underline{13.87}} & \bestfour{\underline{15.52}} & \underline{12.52} & \bestfour{\underline{13.52}}& \bestfour{\underline{14.56}} & \bestfour{\underline{12.54}} & \bestfour{\underline{13.56}} & \bestthree{\underline{14.34}} & \bestfour{\underline{16.24}} & \underline{13.22} & \bestfour{\underline{12.63}} \\
     \multicolumn{2}{l}{DepthSplat~\cite{depthsplat}} & \besttwo{\underline{15.67}} & \besttwo{\underline{16.52}} & \best{\underline{20.90}} & \besttwo{\underline{19.24}} & \besttwo{27.44} & \best{\underline{21.87}} & \bestfour{\underline{22.54}}   & \besttwo{\underline{12.07}} & \bestthree{\underline{12.62}} & \besttwo{\underline{14.15}} & \bestthree{\underline{16.24}} & \bestfour{\underline{13.23}} & \bestthree{\underline{13.77}}& \bestthree{\underline{15.93}} & \besttwo{\underline{14.23}} & \bestthree{\underline{14.01}} & \besttwo{\underline{15.72}} & \bestthree{\underline{16.78}} & \bestthree{\underline{14.35}} & \bestthree{\underline{13.12}} \\
     \multicolumn{2}{l}{LVSM~\cite{jin2024lvsmlargeviewsynthesis}} & - & - & - & - & \best{29.67} & - & - & - & - & -& - & -& - & -& - & -& - & - & - & - \\
    \midrule
    \multicolumn{20}{l}{\textbf{Diffusion-based models}} \\
     \multicolumn{2}{l}{MotionCtrl~\cite{chen2023motion}} & - & - & 12.74 & 16.29 & -  & - & - & - & - & - & - & \bestthree{15.46} & - & - & - & - & - & - & \bestfour{13.29} & - \\
     \multicolumn{2}{l}{4DiM~\cite{4dim}} & - & - & 17.08 & - & - & - & - & - & - & - & -& - & - & -& - & - & - & - & - & - \\
     \multicolumn{2}{l}{ViewCrafter~\cite{yu2024viewcrafter}} & \bestfour{\underline{14.64}} & \bestthree{\underline{15.93}} & \besttwo{\underline{20.43}} & \best{22.04} & \underline{21.42} & \bestthree{\underline{20.88}} & \bestthree{\underline{22.81}} & \bestfour{\underline{10.53}} & \besttwo{\underline{13.52}} & \bestfour{\underline{12.66}} & \besttwo{\underline{16.40}} & \best{18.96} & \besttwo{\underline{14.72}} & \besttwo{\underline{16.42}} & \bestthree{\underline{12.66}} & \besttwo{\underline{14.59}} & \bestfour{\underline{13.78}} & - & \best{18.07} & - \\
     \multicolumn{2}{l}{\ours} & \best{30.30} & \best{31.53} & \bestfour{17.99} & \bestfour{18.56} & \bestfour{25.66} & \bestfour{18.11} & \best{27.57} & \best{14.03}  & \best{19.48} &  \best{14.47} & \best{20.82} & \besttwo{18.40} & \best{19.25} & \best{19.75} & \best{18.91} & \best{16.70} & \best{17.80} & \besttwo{20.96} & \besttwo{15.16} & \besttwo{16.50} \\
    \bottomrule
\end{tabular}
\caption{
\textbf{PSNR$\uparrow$ on small-viewpoint set NVS.}
$P$ denotes the number of input views.
For all results with $P=1$, we sweep the unit length for camera normalization due to the model's scale ambiguity.
O$_e$ and O$_h$ denote the easy and hard split of our split, respectively.
\underline{Underlined} numbers are run by us using the offical released code.
}
\label{tab:setnvs_small}
\end{table*}

\begin{table*}[ht!]
    \centering
    \begin{minipage}{0.63\textwidth}
        \centering
        \tablestyle{0.5pt}{1.0}
        \begin{tabular}{lccccccccccccccc}
            \toprule
            \multirow{3}{*}{Method} & dataset & OO3D & GSO & CO3D & \multicolumn{2}{c}{WRGBD} & \multicolumn{2}{c}{Mip360} & \multicolumn{2}{c}{DL3DV} & \multicolumn{4}{c}{T\&T} \\
            \cmidrule(lr){2-2} \cmidrule(lr){3-3} \cmidrule(lr){4-4} \cmidrule(lr){5-5} \cmidrule(lr){6-7} \cmidrule(lr){8-9} \cmidrule(lr){10-11} \cmidrule(lr){12-15}
             & split & S~\cite{voleti2024sv3d} & S~\cite{voleti2024sv3d} & R~\cite{reconfusion} & \multicolumn{2}{c}{O$_\mathrm{h}$} & \multicolumn{2}{c}{R~\cite{reconfusion}} & \multicolumn{2}{c}{O} & \multicolumn{4}{c}{O} \\
             \cmidrule(lr){2-2} \cmidrule(lr){3-3} \cmidrule(lr){4-4} \cmidrule(lr){5-5} \cmidrule(lr){6-7} \cmidrule(lr){8-9} \cmidrule(lr){10-11} \cmidrule(lr){12-15}
             & $P$ & 1 & 1 & 1 & 1 & 3 & 1 & 3 & 1 & 3 & 1 & 3 & 6 & 9 \\
            \midrule
            \multicolumn{2}{l}{SV3D~\cite{voleti2024sv3d}} & \best{19.28} & \besttwo{20.38} & - & - & - & - & - & - & - & - & - & - & -\\
             \multicolumn{2}{l}{DepthSplat~\cite{depthsplat}} & \bestthree{\underline{11.56}} & \bestthree{\underline{12.32}} & \besttwo{\underline{10.42}} & \besttwo{\underline{9.35}} & \besttwo{\underline{13.53}} & \besttwo{\underline{10.49}} & \bestthree{\underline{12.54}} & \besttwo{\underline{9.63}} & \besttwo{\underline{12.52}} & \bestthree{\underline{8.63}} & \bestthree{\underline{9.78}} & \bestthree{\underline{10.12}} & \besttwo{\underline{11.20}} \\
             \multicolumn{2}{l}{CAT3D~\cite{cat3d}} & - & - & - & - & - & - & \besttwo{15.15} & - & - & - & - & - & -\\
             \multicolumn{2}{l}{ViewCrafter~\cite{yu2024viewcrafter}} & \bestfour{\underline{10.56}} & \bestfour{\underline{11.42}} & \bestthree{\underline{10.11}} & \bestthree{\underline{9.12}} & \bestthree{\underline{13.45}} & \bestthree{\underline{9.79}} & \bestfour{\underline{10.34}} & \bestthree{\underline{8.97}} & \bestthree{\underline{11.50}} & \besttwo{\underline{9.23}} & \besttwo{\underline{9.88}} & \besttwo{\underline{10.32}} & \bestthree{\underline{11.08}} \\
             \multicolumn{2}{l}{\ours} & \besttwo{19.25} & \best{20.65} & \best{15.30} & \best{14.37} & \best{17.28} & \best{12.93} & \best{15.78} & \best{13.01} & \best{15.95} & \best{11.28} & \best{12.65} & \best{13.80} & \best{14.72} \\
            \bottomrule
        \end{tabular}
        \caption{
        \textbf{PSNR$\uparrow$ on large-viewpoint set NVS.} 
        For all results with $P=1$, we sweep the unit length for camera normalization due to the model's scale ambiguity.
        \underline{Underlined} numbers are run by us using the officially released code.
        }
        \label{tab:setnvs_large}
    \end{minipage}
    \hfill
    \begin{minipage}{0.36\textwidth}
        \centering
        \tablestyle{0.8pt}{1.0}
        \begin{tabular}{lccccc}
        \toprule
        \multirow{2}{*}{Method} & \multicolumn{4}{c}{small-viewpoint} & \makecell{large-\\viewpoint} \\
        \cmidrule(lr){2-5} \cmidrule(lr){6-6}
        & RE10K & LLFF & DTU & CO3D & Mip360 \\
        \midrule
        ZipNeRF~\cite{barron2023zip} & \bestfour{20.77}	& \bestfour{17.23} & 9.18 & 14.34 & 12.77 \\
        ZeroNVS~\cite{sargent2023zeronvs} & 19.11	& 15.91 & \bestfour{16.71} & \bestfour{17.13} & \bestfour{14.44} \\
        ReconFusion~\cite{reconfusion} & \bestthree{25.84}	& \bestthree{21.34} & \bestthree{20.74} & \bestthree{19.59} & \bestthree{15.50} \\
        CAT3D~\cite{cat3d} & \besttwo{26.78} & \besttwo{21.58} & \besttwo{22.02} & \besttwo{20.57} &	\besttwo{16.62} \\
        \ours & \best{27.95} & \best{21.88} & \best{22.68} & \best{21.88} & \best{17.82} \\
        \bottomrule
        \end{tabular}
        \caption{
        \textbf{PSNR$\uparrow$ on 3DGS renderings for set NVS.} Results are reported on the ReconFusion~\cite{reconfusion} split with $P=3$.
        }
        \label{tab:3dgs}
    \end{minipage}
    \vspace{-0.5cm}
\end{table*}

\begin{figure*}
    \includegraphics[width=\textwidth]{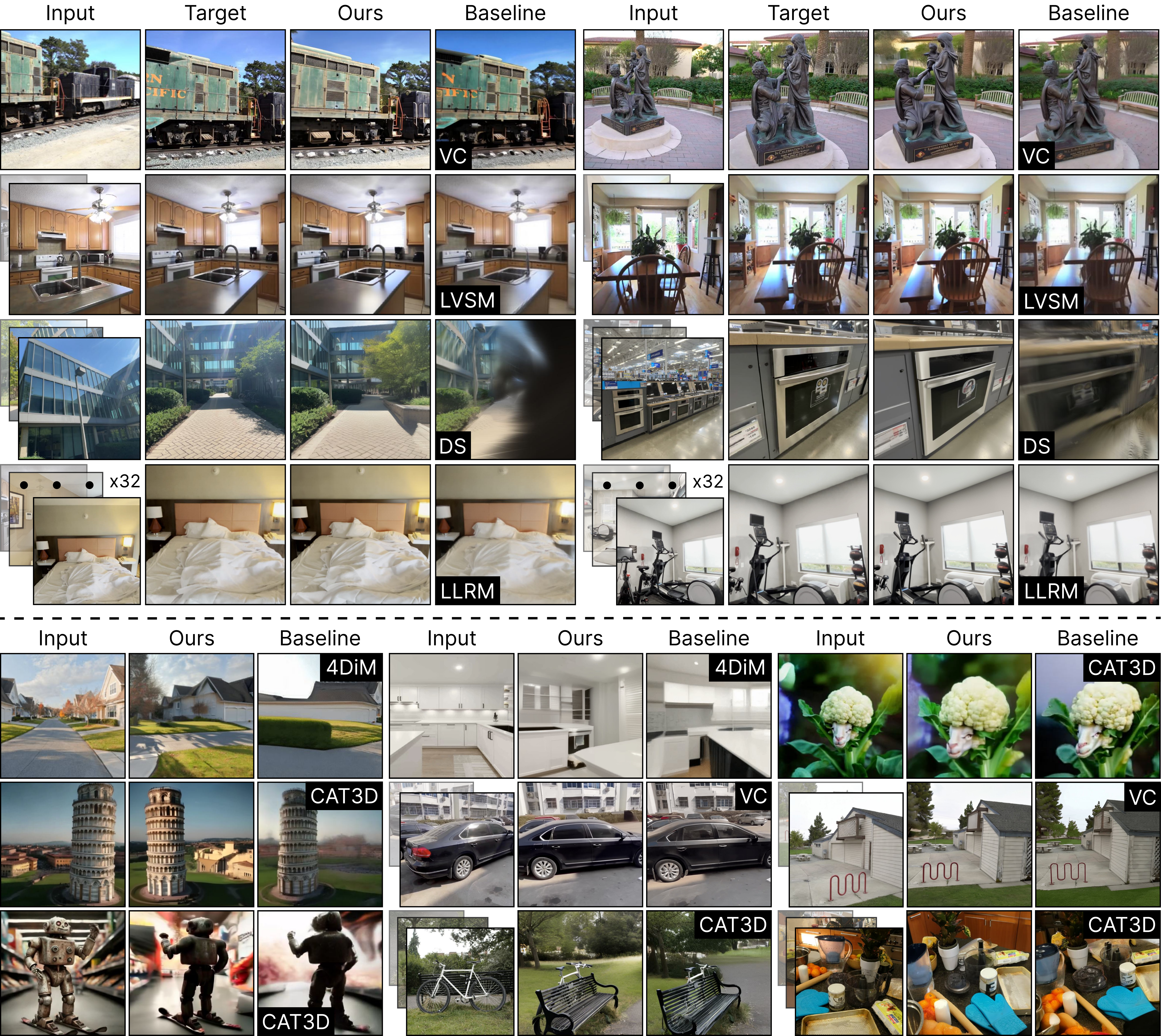}
    \vspace{-0.5cm}
    \caption{
        \textbf{SOTA comparison} on set NVS (top) and trajectory NVS (bottom)  across varying numbers of input views.
        We compare with open-source approaches—ViewCrafter~\cite{yu2024viewcrafter} (VC) and DepthSplat~\cite{depthsplat} (DS)—as well as proprietary ones including LVSM~\cite{jin2024lvsmlargeviewsynthesis}, Long-LRM~\cite{longlrm} (LLRM), 4DiM~\cite{4dim}, and CAT3D~\cite{cat3d}.
        When the input comprises multiple views, we arrange them so that the view closest to the target is placed at the top of each set.
    }
    \label{fig:qualitative}
    \vspace{-0.5cm}
\end{figure*}

In this section, we focus on comparing our model against prior works, given that set NVS is a task that has been extensively explored.

\vspace{-0.3cm}
\paragraph{Quantitative comparison.} 
The input and target views are chosen following splits used in previous methods. The order of target views is not preserved, \ie, $\rmI^\text{tgt} \sim \gI$.
We use standard metrics of peak signal-to-noise ratio (PSNR), learned perceptual image patch similarity~\cite{zhang2018lpips} (LPIPS), and structural similarity index measure~\cite{wang2004image} (SSIM). Only PSNR is showcased here due to space limits, with the rest deferred to~\cref{app:quantitative}. 
Empirically, our method shows a greater performance improvement on LPIPS, reflecting the photorealism of our results.

For small-viewpoint set NVS, \cref{tab:setnvs_small} shows that \ours sets state-of-the-art results in the majority of splits. 
In the sparse-view regime (\ie, $P\leqslant8$), \ours excels across different datasets when $P>1$. For example, a performance gain of +6.0 dB PSNR is achieved on LLFF with $P=3$.
In the semi-dense-view regime (\eg, $P=32$), \ours surprisingly performs favorably against the specialized model~\cite{longlrm}, despite not being specifically designed for this setup. For example, \ours lags behind the state-of-the-art method~\cite{longlrm} by only 1.7 dB on T\&T. 
On object datasets OO3D and GSO, \ours achieves a significantly higher state-of-the-art PSNR compared to all other methods.

Notably, for small-viewpoint set NVS on the RealEstate10K~\cite{realestate10k} dataset, 
\ours underperforms when in the single-view regime (\ie, $P=1$). 
This issue arises from scale ambiguity in the model due to two factors: (1) it always takes in unit-normalized cameras during training, and (2) it is trained on multiple datasets with diverse scales. This challenge is most pronounced on RE10K, where panning motion dominates.
Additionally, the absence of a second input view negates any scale relativity. 
To address this, for all results with $P=1$, we sweep the \textit{unit length for camera normalization} from $0.1$ to $2.0$ (with $2.0$ used during training), selecting the best scale for each scene.
On P split with $P=2$, we observe diffusion models lag behind regression-based models that are advantageous in small-viewpoint interpolation. \ours bridges this gap by improving upon the state-of-the-art diffusion-based model by +4.2 dB.
On R split with $P=3$, the advantage of \ours is pronounced exceeding the previously best result by +1.9 dB.
Notably, ViewCrafter excels on V split due to capacity taking in wide-aspect-ratio images and thus more input pixels than others with square images. The advantage of ViewCrafter on V split diminishes on CO3D since the majority of informative pixels are centrally located. 

For large-viewpoint set NVS, \cref{tab:setnvs_large} shows that \ours's quantitative advantages are even more prominent here, revealing clear benefits of \ours in terms of generation capacity when the camera spans a large spatial range. 
On Mip360 with $P=3$, \ours improves over previous state-of-the-art method CAT3D~\cite{cat3d} by +0.6 dB PSNR. 
On harder scenes like DL3DV and T\&T with different input-view configurations, \ours obtains a clear performance lead. 
On OO3D and GSO with $P=1$, although the performances of \ours and previous state-of-the-art method~\cite{voleti2024sv3d} are similar, we qualitatively observe more photorealistic and sharper output from our model.

\vspace{-0.3cm}
\paragraph{Qualitative comparison.}
\cref{fig:qualitative} top panel shows a qualitative comparison with diverse baselines.
For small-viewpoint set NVS, the output from \ours with the best scale exhibits desirable alignment with the ground truth while being more photorealistic in details. 
Compared with LVSM~\cite{jin2024lvsmlargeviewsynthesis} on the P split of RE10K, \ours produces sharper images, also corroborating that lower PSNR arises from scale ambiguity rather than interpolation quality. 
Similar trends hold when compared to Long-LRM~\cite{longlrm} on DL3DV with $P=32$.
For large-viewpoint set NVS, we compare with DepthSplat~\cite{depthsplat} on DL3DV with $P=3$. 
DepthSplat fails to produce reasonable results when the viewpoint change is too large and falls short in overall visual quality.

\vspace{-0.3cm}
\paragraph{Comparison of 3D reconstruction.} 
To enable a direct quantitative comparison with prior works~\cite{reconfusion,cat3d}, we adopt the few-view 3D reconstruction pipeline described in~\cite{cat3d}. 
For each scene, we first generate 8 videos conditioned on the same input views following different camera paths, summing into 720 generated views.
Then, both the input views and generated views are distilled into a 3DGS-MCMC~\cite{kheradmand20243d} representation without point cloud initialization.
We optimize the camera parameters and apply LPIPS loss~\cite{zhang2018unreasonable} during the distillation.
Finally, we render the distilled 3D model on the test views and report the performance in~\cref{tab:3dgs}.
\ours shows a consistent performance lead.

\subsection{Trajectory NVS}
\label{sec:exp_trajnvs}

In this section, we focus on qualitative demonstration, given that trajectory NVS is an underexplored task. We then compare against prior arts both qualitatively and quantitatively.

\vspace{-0.3cm}
\paragraph{Qualitative results.}
\cref{fig:fancy} presents qualitative results, illustrating trajectories of varying complexities with different numbers of input views across diverse types, including object-centric scene-level, scene-level, real-world, and text-prompted from image diffusion models~\cite{stablediffusion}, \etc.

In the single-view regime (\ie, $P=1$), we manually craft a set of common camera movements/effects, \eg, look-at 360, spiral, panning, zoom-in, zoom-out, dolly zoom, \etc. 
We observe that \ours generalizes to a wide range of images and demonstrates accurate camera-following capacity. 
Excitingly, our model derives reasonable output with a dolly zoom effect (the second row of~\cref{fig:fancy}).
In the \textsc{fern} scene from the third row of~\cref{fig:fancy}, our model demonstrates its ability to generate plausible outputs even when moving close to or passing through an object—despite never being explicitly trained for such scenarios. This highlights the expressiveness of our model. 
An extensive sweeping of camera movements on 4 types of images is provided in~\cref{fig:1view_obj_prompt,fig:1view_scene_prompt,fig:1view_obj,fig:1view_scene}.

In the sparse-view regime with few input views (\ie, $1<P\leqslant8$), 
we observe that \ours demonstrates strong generalization to in-the-wild real-world images and versatility in adapting to different numbers of input views. 
The output forms a smooth trajectory video with subtle temporal flickering, revealing its capacity to interpolate between views smoothly.
In the last row of~\cref{fig:fancy}, our model generates plausible results at the end of the trajectory---an area unseen in the input observations---demonstrating its strong generation capacity.
In the semi-dense-view regime (\ie, $P>9$), we similarly find that \ours is surprisingly able to produce a smooth trajectory video with minimal artifacts. 
Please check the website for video results. 

\vspace{-0.3cm}
\paragraph{Qualitative comparison.} 
\cref{fig:qualitative} bottom panel presents a qualitative comparison with diverse baselines.
In the single-view regime (\ie, $P=1$), we compare to 4DiM~\cite{4dim} and CAT3D~\cite{cat3d}. 
We observe more photo-realistic and sharper output from our model, especially in the background area for object-centric scenes. 4DiM outputs tend to be cartoonish and over-simplistic, given that the model is only trained on RE10K.
In the sparse-view regime with few input views (\ie, $P=3$), we compare with CAT3D and observe that our model demonstrates more photo-realistic textures, especially in the background. For start-end-view interpolation considered in ViewCrafter~\cite{yu2024viewcrafter} with $P=2$, our model produces smooth transitions across trajectories, although it exhibits slight flickering between adjacent frames, particularly in regions with significant high-frequency detail.

\begin{table}
\tablestyle{1.8pt}{1.0}
\centering
\begin{tabular}{lccccccccc}
    \toprule
    \multirow{4}{*}{Method} & \multirow{2}{*}{split} & \multicolumn{5}{c}{small-viewpoint} & \multicolumn{3}{c}{large-viewpoint} \\
    \cmidrule(lr){3-7} \cmidrule(lr){8-10}
    & & \multicolumn{3}{c}{V~\cite{yu2024viewcrafter}} & \multicolumn{5}{c}{O} \\
    \cmidrule(lr){2-2} \cmidrule(lr){3-5} \cmidrule(lr){6-10}
    & dataset & RE & CO3D & T\&T & RE & DTU & WR & DL & T\&T \\
    \midrule
     \multicolumn{2}{l}{MotionCtrl~\cite{chen2023motion}} & 16.29 & \bestfour{15.46} & 13.29 & - & - & - & - & - \\
     \multicolumn{2}{l}{DepthSplat~\cite{depthsplat}} & \besttwo{\underline{19.24}} & \underline{13.23} & \bestfour{\underline{14.35}}  & \bestfour{\underline{25.23}} & \bestfour{\underline{14.68}} & \bestfour{\underline{12.45}} & \bestfour{\underline{11.32}} & \bestfour{\underline{9.11}} \\
     \multicolumn{2}{l}{ViewCrafter~\cite{yu2024viewcrafter}} & \best{22.04} & \best{18.96} & \best{18.07} & \bestthree{\underline{26.54}} & \bestthree{\underline{18.99}} & \bestthree{\underline{13.44}} & \bestthree{\underline{11.45}} & \bestthree{\underline{9.68}} \\
     \midrule 
     \multicolumn{2}{l}{\ours} & \bestfour{18.56} & \bestthree{18.40} & \besttwo{15.16} & \besttwo{27.34} & \besttwo{19.99} & \besttwo{17.79} & \besttwo{15.76} & \besttwo{11.92} \\
     \multicolumn{2}{l}{\ours (+ temp.)} & \bestthree{18.62} & \besttwo{18.43} & \bestthree{15.13} & \best{27.36} & \best{20.19} & \best{17.93} & \best{15.78} & \best{11.99} \\
    \bottomrule
\end{tabular}
\caption{
\textbf{PSNR$\uparrow$ on trajectory NVS. } 
\textit{temp.} denotes optional temporal pathway.
RE, WR, and DL denotes RE10K, WRGBD, and DL3DV, respectively.
For the V~\cite{yu2024viewcrafter} split, $P=1$ with unit length swept; for the O split, $P=3$. 
\underline{Underlined} numbers are run by us using the officially released code.
}
\label{tab:trajnvs}
\vspace{-0.2cm}
\end{table}

\begin{table}
\tablestyle{4.0pt}{1.0}
\centering
\begin{tabular}{lccccc}
\toprule
\multirow{2}{*}{Method} & \multicolumn{2}{c}{samples}  & 3DGS & video \\
\cmidrule(lr){2-3} \cmidrule(lr){4-4} \cmidrule(lr){5-5}
& PSNR$\uparrow$ & TSED$\downarrow$ & PSNR$\uparrow$ & MS$\uparrow$  \\
\midrule
\ours (one-pass) & \bestthree{15.73} & \besttwo{115.1} & \bestthree{16.03} &  \bestfour{95.39} \\
\ours (two-pass: $\mathrm{nearest}$) & 13.74 & 120.9 & 14.21 &  94.71 \\
\ours (two-pass: $\mathrm{gt+nearest}$) & 15.58 & \bestthree{116.2} & 15.96 & 95.22 \\
\ours (two-pass: $\mathrm{gt+interp}$) & \bestfour{15.66} & 120.1 & \bestfour{15.98} &  \bestthree{95.56} \\
\midrule
\ours & \besttwo{15.76} & \bestfour{116.7} & \besttwo{16.11} & \besttwo{95.76} \\
\ours (+ temp.) & \best{15.78} & \best{109.0} & \best{16.17} &  \best{95.77} \\
\bottomrule
\end{tabular}
\caption{
\textbf{3D consistency (TSED$\downarrow$ and PSNR$\uparrow$) and temporal quality (MS$\uparrow$) on trajectory NVS.}
\ours uses $\mathrm{interp}$ procedural sampling by default. 
\textit{temp.} denotes the optional temporal pathway.
\textit{MS} denotes motion smoothness from VBench~\cite{huang2023vbench}.
Results are reported on our split of DL3DV with $P=3$.
}
\label{tab:trajnvs_video}
\vspace{-0.3cm}
\end{table}

\begin{figure}
    \includegraphics[width=\linewidth]{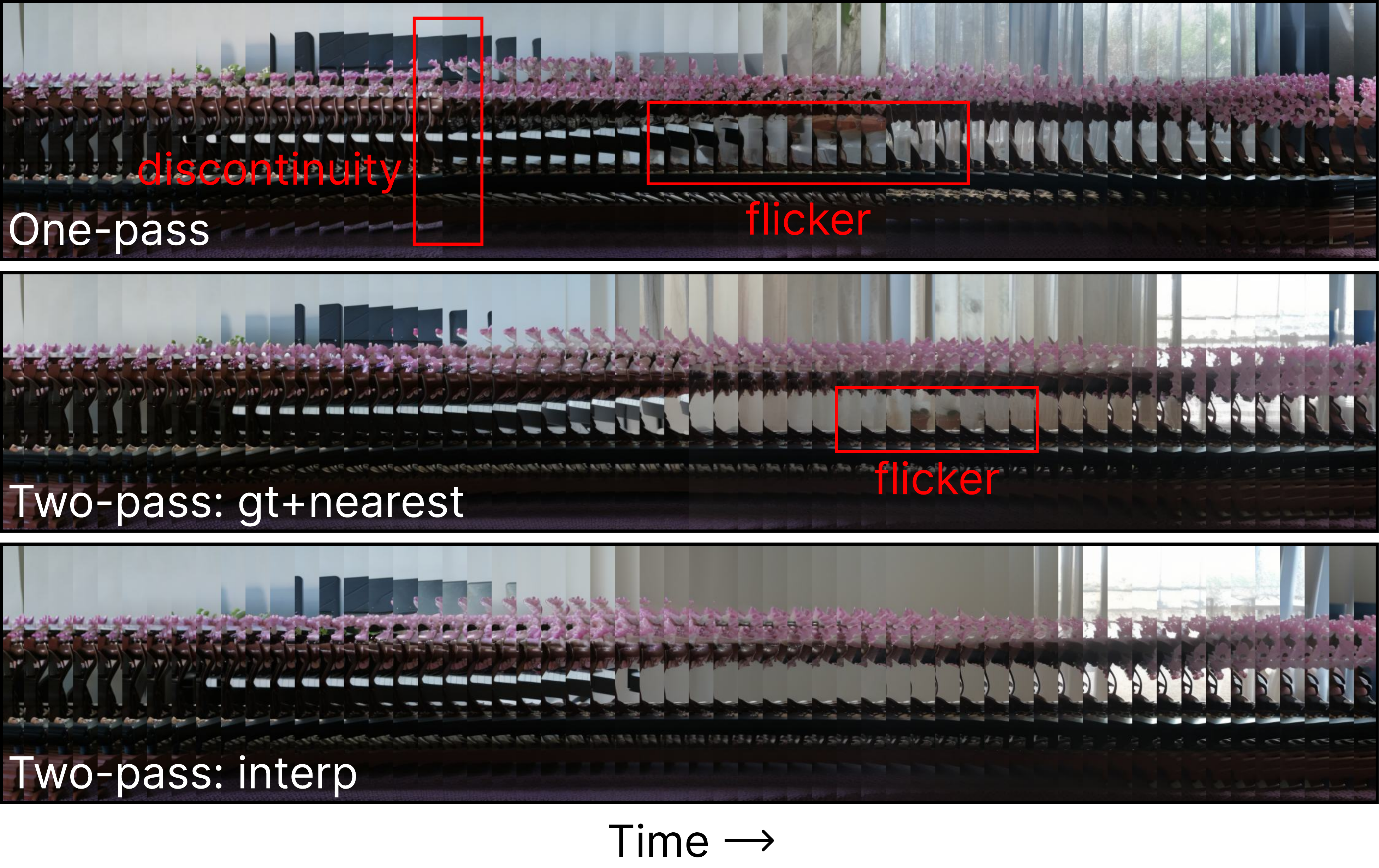}
    \vspace{-0.7cm}
    \caption{
        \textbf{Temporal quality.         
        }
        Vertical slices of a rendered novel camera path on the \textsc{bonsai} scene from Mip-NeRF360~\cite{barron2021mip} illustrate the temporal quality across adjacent viewpoints. One-pass or $\mathrm{gt+nearest}$ procedural sampling results in notable flickering, whereas $\mathrm{interp}$ procedural sampling ensures temporally smooth rendering.
    }
    \label{fig:ablation-temporal-quality}
    \vspace{-0.5cm}
\end{figure}

\vspace{-0.3cm}
\paragraph{Quantitative comparison.} 
We use the same input views as in the set NVS for each split. We use all frames from each scene as target views such that they form a smoothly transitioning trajectory video, \ie, $\rmI^\text{tgt} \sim \gV$. We use PSNR as metrics and compare with baselines in~\cref{tab:trajnvs}. 

For small-viewpoint trajectory NVS, \cref{tab:trajnvs} compares \ours with baselines on PSNR. \ours performs favorably against other methods in V split with $P=1$. 
The performance lead of ViewCrafter is mainly attributed to its training on high-resolution images. 
For large-viewpoint trajectory NVS with $P=3$, \ours consistency sets new state-of-the-art results. 
Applying the temporal pathway further boosts performance and improves smoothness, indicating the benefits of the gated architecture.

\vspace{-0.3cm}
\paragraph{Ablation on two-pass procedural sampling.}
We conduct an ablation study comparing the default $\mathrm{interp}$ procedural sampling with one-pass sampling and alternative procedural sampling strategies.

Quantitatively, beyond evaluating PSNR on individual views, we assess 3D consistency using the PSNR on 3D renderings of that same view and SED~\cite{4dim,yu2023long} score. 
To compute the SED score, we first apply SIFT~\cite{zhou2009object} to detect keypoints in two images. For each keypoint in the first image, we determine its corresponding epipolar line in the second image and measure the shortest distance to its match.
Additionally, we report $\textrm{Motion Smoothness}$ (MS) from VBench~\cite{huang2023vbench}, a benchmark designed to evaluate temporal coherence in video generative models. As shown in~\cref{tab:trajnvs_video}, $\mathrm{interp}$ procedural sampling demonstrates a clear advantage over its alternatives, with the integration of the temporal pathway further reinforcing its superiority.

Qualitative comparisons in~\cref{fig:ablation-temporal-quality} show that one-pass sampling introduces visible temporal flickering and abrupt visual changes. In contrast, $\mathrm{interp}$ produces the smoothest transitions, outperforming $\mathrm{gt+nearest}$ and mitigating noticeable flickering.

\begin{figure}
    \includegraphics[width=\linewidth]{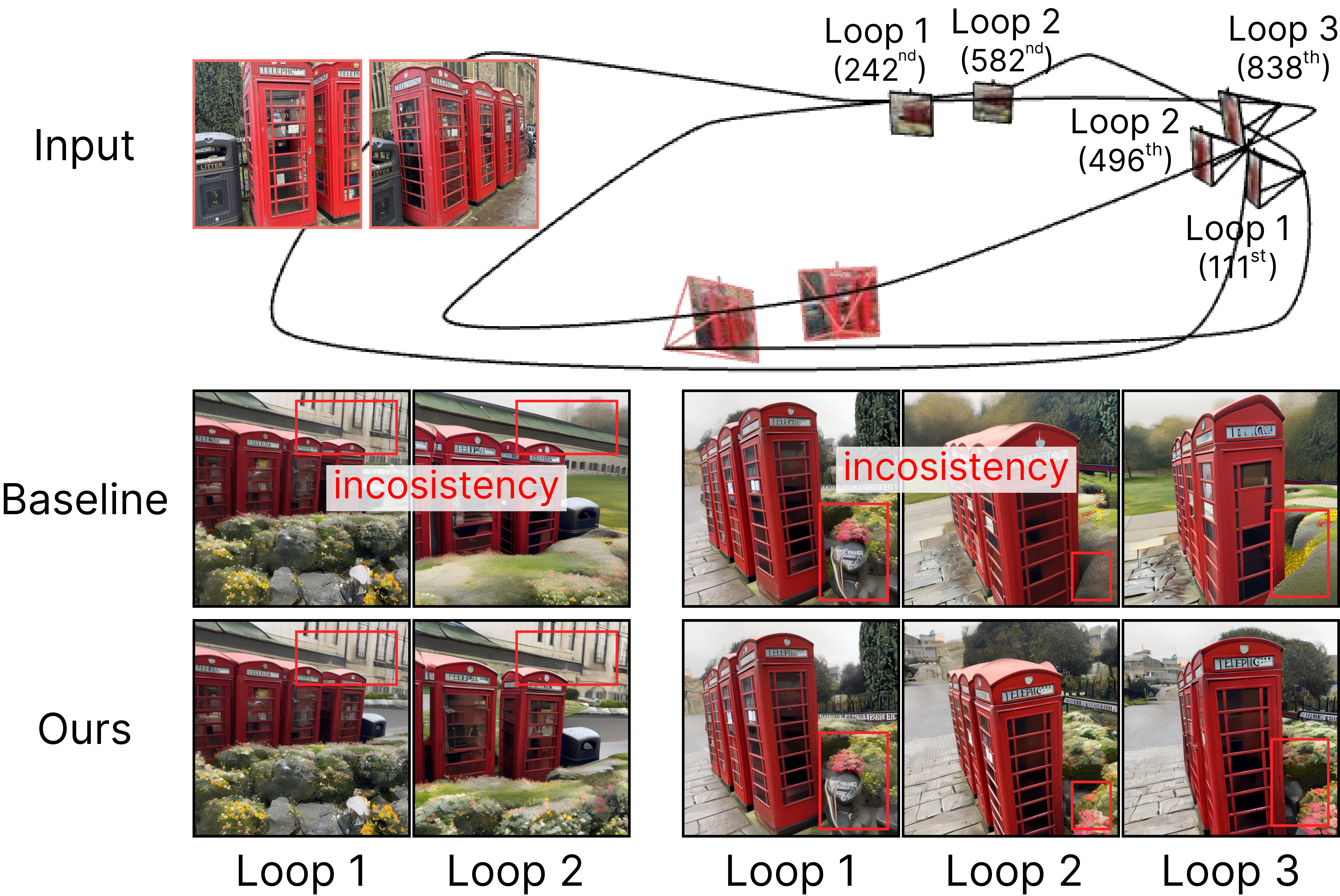}
    \vspace{-0.5cm}
    \caption{
        \textbf{Long-range 3D consistency.}
        We visualize samples following a camera path looping three times around the \textsc{telephone-booth} scene. 
        Lookup using spatial neighbors from the memory bank (ours) notably improves view consistency and reduces artifacts in recurring locations across different loops, compared to lookup using temporal neighbors (baseline).
    }
    \label{fig:ablation-3d-consistency}
    \vspace{-0.5cm}
\end{figure}

\subsection{Long-Trajectory NVS}
\label{sec:exp_trajnvs_long}

\cref{fig:ablation-3d-consistency} presents a qualitative demonstration of NVS over a long trajectory of up to 1000 frames.
As the camera orbits the \textsc{telephone booth} for multiple rounds, the generated views in each round from similar viewpoints can be drastically different since they are far away from each other temporally. 
With the memory bank maintaining previously generated anchors, \ours achieves robust 3D consistency for long-trajectory NVS, \eg, the building in front of and the plantation after the booth. 
Comparing it to using temporal nearest anchors previously generated, using spatially nearest ones demonstrates a clear advantage.
The memory mechanism has been concurrently explored in previous works~\cite{yu2024viewcrafter, Ma2024See3D}, leveraging explicit intermediate 3D representations such as dense point clouds predicted by DUSt3R~\cite{dust3r_cvpr24}. In contrast, our model demonstrates greater robustness and generalizability to in-the-wild data, as it is not constrained by the quality of DUSt3R’s output, which often becomes unreliable in quality for data outside of its training domain, \eg, text-prompted images.

\subsection{Discussions}
\label{sec:discussions}

\begin{figure}
    \includegraphics[width=\linewidth]{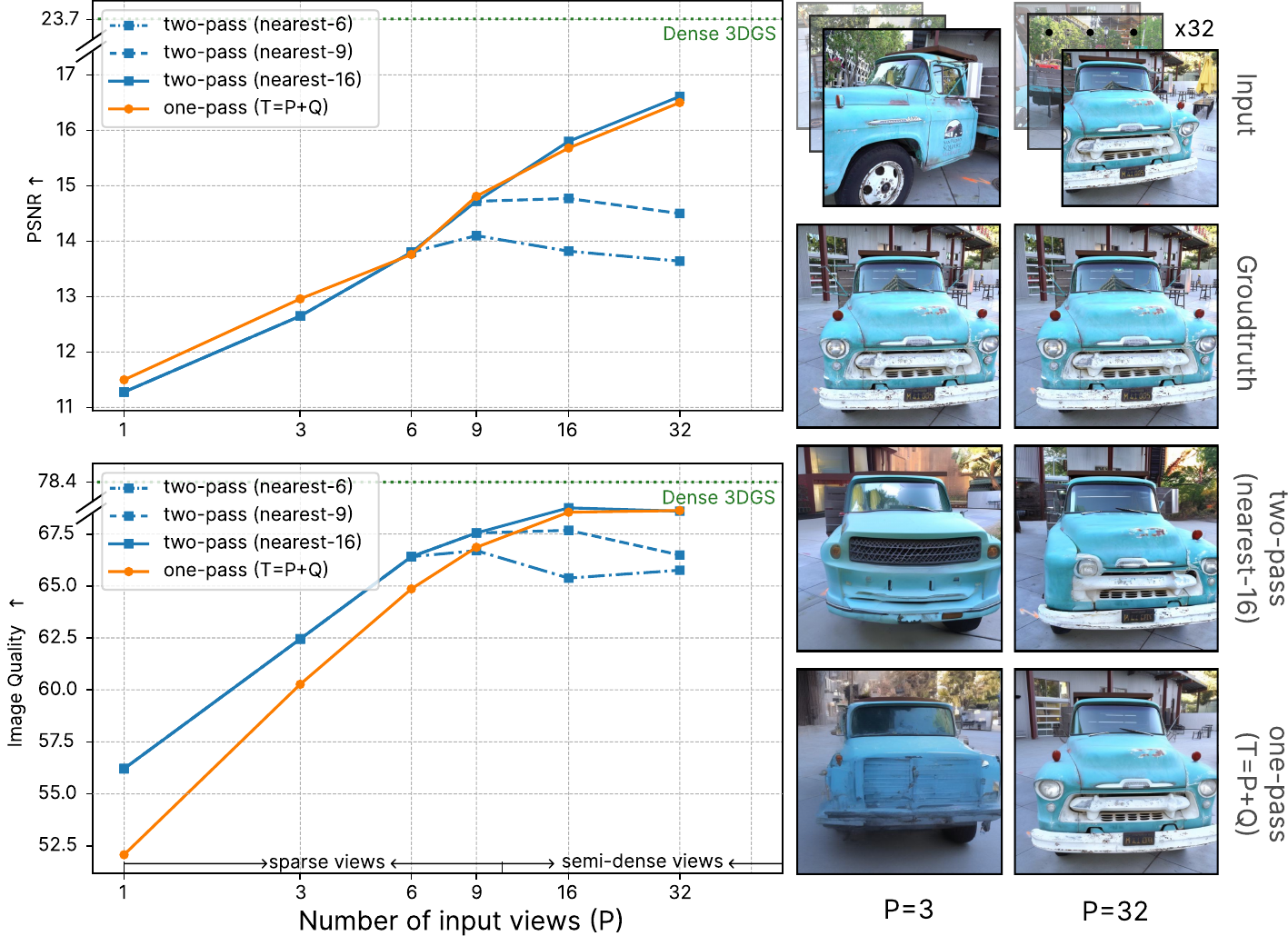}
    \vspace{-0.5cm}
    \caption{
        \textbf{Generation quality on the number of input views.}
        PSNR$\uparrow$ (top) and Image Quality$\uparrow$ (bottom) on set NVS. Results are reported on our split of T\&T. Extending $T$ to more input views in a zero-shot manner produces more consistent samples in the semi-dense-view regime.
        Dense 3DGS denotes results of~\cite{kerbl3Dgaussians} with full views.
    }
    \label{fig:ablation-input-views}
    \vspace{-0.5cm}
\end{figure}

\paragraph{Zero-shot generalization of context window length $T$.}
We surprisingly find our model, though only trained on $T=21$ frames, can generalize reasonably to larger $T$ during sampling \textit{in the semi-dense-view regime}.
On our split of T\&T for set NVS, we evaluate the predictions against ground truth in both sparse-view (\ie, $1\leq P\leq 8$) and semi-dense-view regime (\ie, $9\leq P$) using PSNR$\uparrow$ and Image Quality$\uparrow$~\cite{huang2023vbench}. Image Quality refers to the distortion (\eg, over-exposure, noise, blur) presented in the generated image.
We experiment with different sampling strategies: one-pass sampling zero-shot extending the context window length $T$; 
two-pass procedural sampling by first generating anchor views using $\mathrm{nearest-}K$ ($K<T$) input views and then interpolating anchor views into target views.

Our results are shown in~\cref{fig:ablation-input-views}.
Procedural sampling with the $\mathrm{nearest-}K$ anchor views plateau after taking $K$ views as input, 
indicating inefficiencies in procedural sampling and an inability to effectively utilize all available input views when $P>T$. Conversely, the metrics steadily improve with respect to the number of input frames for one-pass sampling with $T$ extending to $P+Q$ in a zero-shot manner.
However, we observe that this generalization fails in the sparse-view regime, resulting in blurry samples, as indicated by the low Image Quality when $P<9$ and qualitative samples when $P=3$.
In the semi-dense-view setting, although quantitative metrics show minimal differences between one-pass and procedural sampling, we consistently observe that one-pass produces more 3D-consistent samples, as illustrated in the bottom-right figure.

\begin{figure}
    \includegraphics[width=\linewidth]{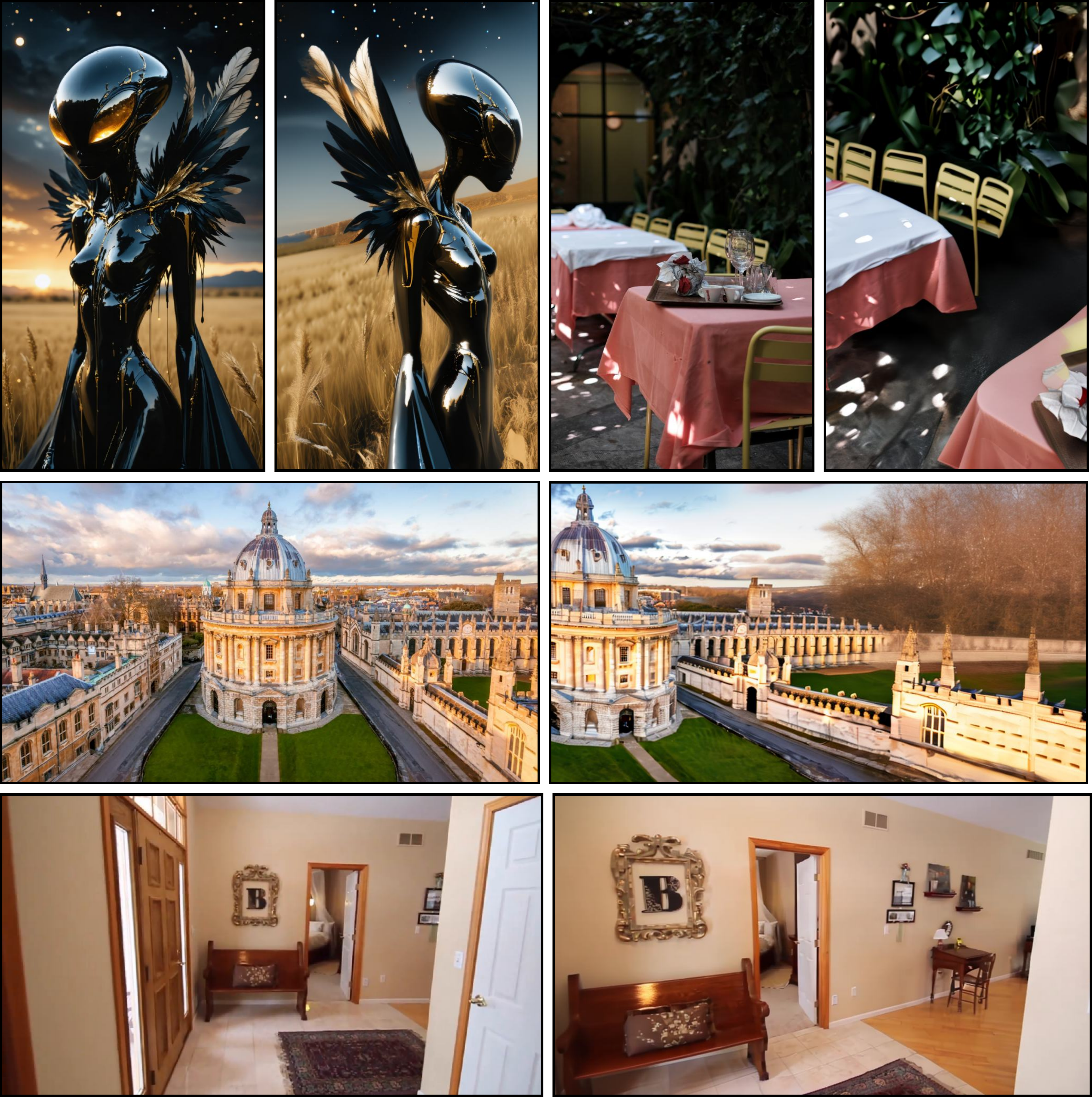}
    \vspace{-0.5cm}
    \caption{
        \textbf{Generation quality on different image resolutions.}
        Our model generalizes to different image resolution of varying aspect ratios, including both portrait (top) and landscape orientations (bottom). 
        Results are presented as a pair of the input view and the target views.
    }
    \label{fig:nonsquare}
\end{figure}

\vspace{-0.3cm}
\paragraph{Zero-shot generalization of image resolution.}
Surprisingly, we find our model, despite being trained only on square images with $H=W=576$, generalizes well to different image resolution during sampling, similar to~\cite{rombach2022high}.
As shown in~\cref{fig:nonsquare}, \ours can produce high-quality results in both portrait ($16:9$) and landscape ($9:16$) orientations of different image resolutions.

\begin{figure}
    \includegraphics[width=\linewidth]{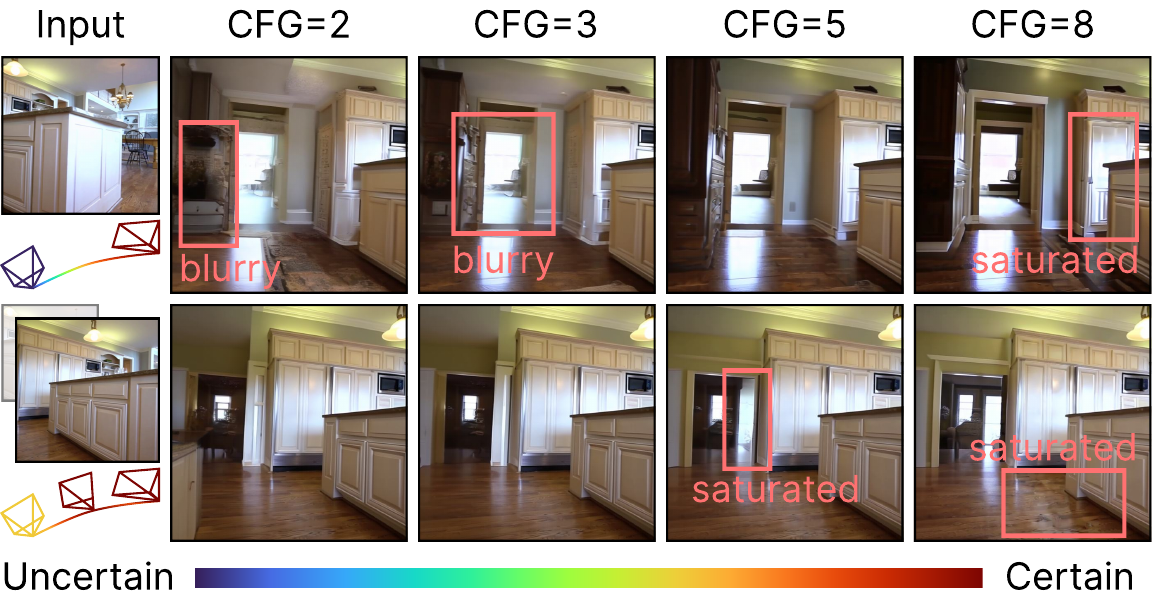}
    \vspace{-0.5cm}
    \caption{
        \textbf{Generation uncertainty on CFG.}
        The CFG scale should be increased as generation uncertainty rises. For single-view conditioning (top), a higher CFG scale is typically required, whereas few-view conditioning (bottom) benefits from a lower scale.
    }
    \label{fig:ablation-cfg}
    \vspace{-0.5cm}
\end{figure}

\vspace{-0.3cm}
\paragraph{Guidance scale on generation uncertainty.}  
We employ classifier-free guidance~\cite{ho2022classifier} (CFG) to enhance sampling quality.  
Empirically, we find that the CFG scale, a hyperparameter at test time, has a significant impact on the final result~\cite{svd}, as shown in~\cref{fig:ablation-cfg}.  
Specifically, the optimal CFG scale is strongly correlated with the inherent uncertainty of the generation.  
When uncertainty is high (top row), a higher CFG scale (\eg, $5$) is preferable to prevent excessive blurriness in the generated samples. 
Conversely, when uncertainty is low (bottom row), a lower CFG scale (\eg, $3$) helps avoid oversaturation.  
In practice, setting the CFG scale between $2$ and $5$ consistently produces high-quality results across all our samples.

\begin{figure}
    \includegraphics[width=\linewidth]{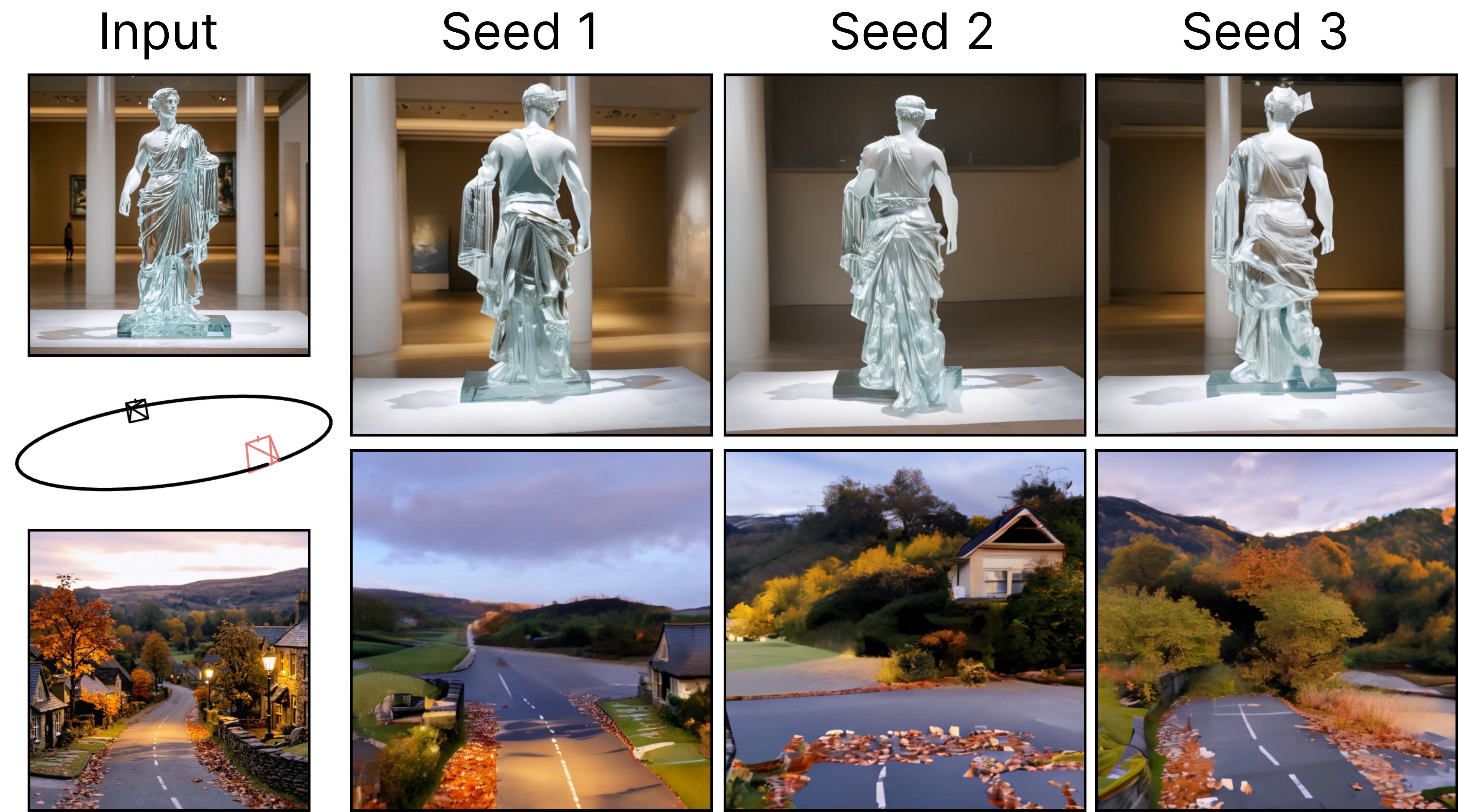}
    \vspace{-0.5cm}
    \caption{
        \textbf{Generation diversity in unseen regions.}
        Our model generates diverse samples by varying randomization seeds during the sampling process.
    }
    \label{fig:diversity}
    \vspace{-0.5cm}
\end{figure}

\vspace{-0.3cm}
\paragraph{Sampling diversity of unseen areas.} 
\cref{fig:diversity} demonstrates the capability of the model to generate diverse and plausible predictions for unseen regions of input observations. 
In the first row, the input view depicts a frontal view of a classical statue. We sample multiple back views by varying the random seeds, producing distinct yet coherent interpretations of the unseen geometry and texture while preserving fidelity to the input. Similarly, in the second row, the model generates multiple plausible continuations of the scene given an input view of a scenic road, each reflecting unique variations in environmental and structural details. 
These results highlight the model's ability to synthesize realistic and diverse outputs for occluded or ambiguous regions.

\section{Conclusion}
\label{sec:conclusion}

We present \textsc{\underline{S}tabl\underline{e} \underline{V}irtual C\underline{a}mera} (\ours), a generalist diffusion model for novel view synthesis that balances large viewpoint changes and smooth interpolation while supporting flexible input and target configurations.
By designing a diffusion-based architecture without 3D representation, a structured training strategy, and a two-pass procedural sampling approach, \ours achieves 3D consistent rendering across diverse NVS tasks.
Extensive benchmarking demonstrates its superiority over existing methods, with strong generalization to real-world scenes. For broader impact and limitations, please refer to the appendix.

\vspace{0.7cm}
\noindent \textbf{Acknowledgment.} We would like to thank Hongsuk Benjamin Choi, Angjoo Kanazawa, Ethan Weber, Ruilong Li, Brent Yi, Justin Kerr, Rundi Wu, Jianyuan Wang, Zihang Lai, Ruining Li, and Gabrijel Boduljak for their thoughtful feedback and discussion. We would like to thank Wangbo Yu, Aleksander Hołyński, Saurabh Saxena, and Ziwen Chen for their kind clarification on experiment settings. We would like to thank Jan-Niklas Dihlmann, Fei Yin, Andreas Engelhardt, and Emmanuelle Bourigault for sharing their amazing phone captures.

\clearpage

{
    \small
    \bibliographystyle{unsrt}
    \bibliography{main}
}

\clearpage
\setcounter{page}{1}

\appendix

\begin{center}
\begin{minipage}{0.98\textwidth} %
    \vspace{-1.0cm}
    \footnotesize
    \setcounter{tocdepth}{4}
    \tableofcontents
\end{minipage}
\end{center}
\clearpage

\section{Broader Impact and Limitations}
\label{app:impactandlimit}

\paragraph{Broader Impact.}
\ours significantly advances immersive 3D experiences by synthesizing realistic and temporally consistent views from sparse camera inputs, addressing key limitations in NVS. Inspired by James Cameron's pioneering Virtual Camera technology—which enabled filmmakers to intuitively navigate virtual environments and visualize precise camera trajectories—our generative AI-driven model similarly allows users to create intricate, controllable camera paths without the typical complexity of dense view captures or explicit 3D reconstructions. By generalizing across arbitrary viewpoint changes and enabling temporally smooth rendering without NeRF distillation, our approach simplifies the NVS pipeline, enhancing accessibility for content creators, developers, and researchers. This facilitates applications ranging from virtual cinematography and gaming to digital heritage preservation, substantially broadening the usability and scalability of NVS.

\paragraph{Limitations.} 
The performance of \ours is constrained by the scope of its training data, resulting in reduced quality for certain types of scenes. Specifically, input images featuring humans, animals, or dynamic textures (\eg, water surfaces) typically lead to degraded outputs. Additionally, highly ambiguous scenes or complex camera trajectories pose challenges; for instance, trajectories that intersect with objects or surfaces may cause noticeable flickering artifacts. Similar issues arise for extremely irregularly shaped objects or when target viewpoints significantly diverge from the provided input viewpoints.

\section{Related Work}
\label{sec:related}

\paragraph{Novel view synthesis.}
While traditional NVS has been studied for nearly several decades,
it has recently achieved remarkable success with the help of techniques such as NeRF~\cite{mildenhall2020nerf,metzer2023latentnerf} and diffusion models~\cite{ho2020denoising,song2019generative}. Using these techniques, there are broadly two ways of generating novel views : 1) estimate a 3D representation using multiple sparse input views, then regress the novel views from this intermediate representation, 2) directly estimate the novel views from the sparse input views, either in a single shot in a feed-forward manner, or in multiple sampling steps using diffusion models.

\paragraph{Feed-forward models.}
Approaches like LFNR~\cite{suhail2022light} and LVSM~\cite{jin2024lvsmlargeviewsynthesis} directly generate target views and leverage data-driven learning to capture 3D inductive biases.
While often efficient, these methods struggle with the inherent diversity of generative NVS, limiting their capacity to model multiple plausible solutions.
In contrast, our approach frames generative NVS through a diffusion perspective, enabling us to sample diverse, plausible solutions during inference, thereby addressing ambiguities and enhancing generation capacity.

\paragraph{Intermediate representation models.}
Techniques such as NeRF~\cite{mildenhall2020nerf} and Gaussian Splatting~\cite{kerbl3Dgaussians} have made significant progress on per-scene optimization from input views by building 3D representations efficiently. Several works show that these representations can then be used to regress novel views. pixelNeRF~\cite{yu2020pixelnerf} builds a NeRF from multiple input views; Splatter Image~\cite{szymanowicz2024splatter}, pixelSplat~\cite{charatan23pixelsplat}, and MVSplat~\cite{chen2025mvsplat} build a 3D representation using Gaussian Splatting; LRM~\cite{hong2023lrm} builds a triplane representation. 
However, these optimization-based methods cannot creatively synthesize missing regions, and rely on tens, if not hundreds, of posed input images which limits their practicality in real-world applications.

\paragraph{Diffusion-based models.}
Our work falls within this category, where target novel views are generated in multiple steps through a denoising diffusion process~\cite{ho2020denoising,song2019generative}.
As mentioned earlier, existing diffusion-based methods can be divided into two main types: \textit{image models} and \textit{video models}.

Image models are designed to synthesize distant viewpoints~\cite{watson2022novel,liu2023zero1to3,shi2023zero123++}. However, these early practices only generate one viewpoint at a time, and lack multi-view consistency, often resulting in jittery and inconsistent samples when generating along a camera trajectory. Works such as MVDream~\cite{shi2023mvdream},  SyncDreamer~\cite{liu2023syncdreamer} and HexGen3D~\cite{mercier2024hexagen3d} generate multiple fixed views simultaneously. 
However, these models only generate specific views given a conditional image, not arbitrary viewpoints.

To obtain consistent 3D objects, these models necessitate NeRF distillation, either through Score Distillation Samling (SDS)~\cite{poole2022dreamfusion,sargent2023zeronvs} or directly upon completely sampled images~\cite{reconfusion,cat3d}. 

Video models can produce smooth video sequences by maintaining certain constraints relative to the input views~\cite{voleti2024sv3d}. However, they are generally limited to smaller camera motions due to the natural frame rate in video training.
Some works use video diffusion models to generate 4D scenes~\cite{bahmani20234d}. But in those works, the video diffusion models do not contribute to the consistency of the 3D object itself, that part is handled by image-based diffusion models such as MVDream.

\section{Benchmark}
\label{app:benchmark}

\begin{table*}
    \tablestyle{2.5pt}{1.0}
    \centering
    \begin{tabular}{lcccccc}
        \toprule
        & \textbf{type} & \textbf{split} & \textbf{\#scene}  & \textbf{$(\rmI^\text{inp}, \rmI^\text{tgt}) \sim \gV$} & \textbf{$P$} & $\mathcal{D}_\mathrm{CLIP}(\rmI)$ \\
        \midrule\midrule
        \multicolumn{7}{l}{\textbf{Small-viewpoint NVS}} \\
        \rule{0pt}{10pt}
        OmniObject3D~\cite{wu2023omniobject3d} & \faCube & O (dynamic orbit) & 308 & \cmark & 3 & 0.11 \\
        \cmidrule(lr){1-1}
        GSO~\cite{downs2022google} & \faCube & O (dynamic orbit)  & 300 & \cmark & 3  & 0.11 \\
        \cmidrule(lr){1-1}
        \multirow{5}{*}{RealEstate10K~\cite{realestate10k}} & \multirow{4}{*}{\faMountain} & D~\cite{4dim} & 128 & \cmark & 1  & 0.09 \\
        & & \multirow{2}{*}{R~\cite{reconfusion}} & \multirow{2}{*}{10} & \multirow{2}{*}{\cmark} & 1 & 0.08 \\
        & & & & & 3 & 0.03 \\
        & & P~\cite{charatan23pixelsplat} & 6474  & \cmark & 2 & 0.04 \\
        & & V~\cite{yu2024viewcrafter} & 10 & \cmark & 2 & 0.11 \\
         \cmidrule(lr){1-1}
        \multirow{2}{*}{LLFF~\cite{mildenhall2019local}} & \multirow{2}{*}{\faMountain} & \multirow{2}{*}{R~\cite{reconfusion}} & \multirow{2}{*}{8} & \multirow{2}{*}{\cmark} & 1  & 0.04 \\
        & & &  & & 3 & 0.03 \\
        \cmidrule(lr){1-1}
        \multirow{2}{*}{DTU~\cite{jensen2014large}} & \multirow{2}{*}{\faMountain} & \multirow{2}{*}{R~\cite{reconfusion}} & \multirow{2}{*}{15} & \multirow{2}{*}{\cmark} & 1 & 0.07 \\
        & & &  & & 3 & 0.06 \\
        \cmidrule(lr){1-1}
        \multirow{2}{*}{CO3D~\cite{co3d}} & \multirow{2}{*}{\faMountain} & R~\cite{reconfusion} & 20 &  \cmark & 3 & 0.09 \\
        & & V~\cite{yu2024viewcrafter} & 10 &  \cmark & 2 & 0.09 \\
        \cmidrule(lr){1-1}
        \multirow{2}{*}{WildRGB-D~\cite{wildrgbd}} & \multirow{2}{*}{\faMountain} & O$_e$ ($1/3$ orbit) & \multirow{2}{*}{20} & \multirow{2}{*}{\cmark} & 3 & 0.07 \\
        & & O$_h$ (full orbit) & & & 6 & 0.11 \\
        \cmidrule(lr){1-1}
        Mip-NeRF360~\cite{barron2021mip} & \faMountain & R~\cite{reconfusion} & 9 & \xmark & 6 & 0.11 \\
        \cmidrule(lr){1-1}
        \multirow{2}{*}{DL3DV-140~\cite{dl3dv10k}} & \multirow{2}{*}{\faMountain} & O & 10 & \cmark & 6 & 0.10 \\
        & & L~\cite{longlrm} & 140  & \cmark & 32 & 0.05 \\
        \cmidrule(lr){1-1}
        \multirow{2}{*}{Tanks and Temples~\cite{knapitsch2017tanks}} & \multirow{2}{*}{\faMountain} & V~\cite{yu2024viewcrafter} & 22  & \cmark & 2 & 0.10 \\
         & & L~\cite{longlrm} & 2 & \cmark & 32 & 0.10 \\
        \midrule
        \multicolumn{5}{l}{\textbf{Large-viewpoint NVS}} \\
        \rule{0pt}{10pt}
        OmniObject3D~\cite{wu2023omniobject3d} & \faCube & S~\cite{voleti2024sv3d} (dynamic orbit) & 308 & \cmark & 1 & 0.16 \\
        \cmidrule(lr){1-1}
        GSO~\cite{downs2022google} & \faCube & S~\cite{voleti2024sv3d} (dynamic orbit) & 300 & \cmark & 1 & 0.18 \\
        \cmidrule(lr){1-1}
        CO3D~\cite{co3d} & \faMountain & R~\cite{reconfusion} & 20 & \cmark & 1 & 0.15 \\
        \cmidrule(lr){1-1}
        \multirow{2}{*}{WildRGB-D~\cite{wildrgbd}} & \multirow{2}{*}{\faMountain} & \multirow{2}{*}{O$_h$ (full orbit)} & \multirow{2}{*}{20} & \multirow{2}{*}{\cmark} & 1 & 0.19 \\
        & & & &  & 3 & 0.14 \\
        \cmidrule(lr){1-1}
        \multirow{2}{*}{Mip-NeRF360~\cite{barron2021mip}} & \multirow{2}{*}{\faMountain} & \multirow{2}{*}{R~\cite{reconfusion}} & \multirow{2}{*}{9} & \multirow{2}{*}{\xmark} & 1 & 0.19 \\
        & & & & & 3 & 0.13 \\
        \cmidrule(lr){1-1}
       \multirow{2}{*}{DL3DV-140~\cite{dl3dv10k}} & \multirow{2}{*}{\faMountain} & \multirow{2}{*}{O} & \multirow{2}{*}{10} & \multirow{2}{*}{\cmark} & 1 & 0.21 \\
        & & &  & & 3 & 0.12 \\
        \cmidrule(lr){1-1}
        \multirow{4}{*}{Tanks and Temples~\cite{knapitsch2017tanks}} & \multirow{4}{*}{\faMountain} & \multirow{4}{*}{O} & \multirow{4}{*}{2} & \multirow{4}{*}{\cmark} & 1 & 0.21 \\
        & & &  & & 3 & 0.18 \\
        & & &  & & 6 & 0.16 \\
        & & &  & & 9 & 0.14 \\
        \bottomrule
    \end{tabular}
    \caption{
    \textbf{Statistics for NVS benchmark}.
    We consider $10$ publicly available datasets commonly used for evaluating NVS, encompassing both object-level and scene-level data.
    Views from Mip-NeRF360~\cite{barron2021mip} derive from several disjoint captures following different camera trajectories, thus all views $(\rmI^\text{inp}, \rmI^\text{tgt}) \sim \gI$.
    $P$ denotes the number of input views.
    Depending on the disparity between $\rmI^\text{inp}$ and $\rmI^\text{tgt}$, we group NVS tasks into small-viewpoint NVS (top panel) where target views are similar to input views and large-viewpoint NVS (bottom panel) where target views are more different to input views.
    }
    \label{tab:benchmark_dataset}
\end{table*}

\begin{table*}
\tablestyle{1.2pt}{1.0}
\centering
\subfloat[\textbf{LPIPS$\downarrow$}]{
    \begin{tabular}{lccccccccccccccccccccc}
    \toprule
    \multirow{3}{*}{Method} & dataset & OO3D & GSO & \multicolumn{5}{c}{RE10K} & \multicolumn{2}{c}{LLFF} & \multicolumn{2}{c}{DTU} & \multicolumn{2}{c}{CO3D} & \multicolumn{2}{c}{WRGBD} & Mip360 & \multicolumn{2}{c}{DL3DV} & \multicolumn{2}{c}{T\&T} \\
    \cmidrule(lr){2-2} \cmidrule(lr){3-3} \cmidrule(lr){4-4} \cmidrule(lr){5-9} \cmidrule(lr){10-11} \cmidrule(lr){12-13} \cmidrule(lr){14-15} \cmidrule(lr){16-17} \cmidrule(lr){18-18} \cmidrule(lr){19-20} \cmidrule(lr){21-22}
     & split & O & O & D~\cite{4dim} & V~\cite{yu2024viewcrafter} & P~\cite{charatan23pixelsplat} & \multicolumn{2}{c}{R~\cite{reconfusion}}   & \multicolumn{2}{c}{R~\cite{reconfusion}} & \multicolumn{2}{c}{R~\cite{reconfusion}} & V~\cite{yu2024viewcrafter} & R~\cite{reconfusion} & O$_\mathrm{e}$ & O$_\mathrm{h}$ & R~\cite{reconfusion} & O & L~\cite{longlrm} & V~\cite{yu2024viewcrafter} & L~\cite{longlrm} \\
     \cmidrule(lr){2-2} \cmidrule(lr){3-3} \cmidrule(lr){4-4} \cmidrule(lr){5-5} \cmidrule(lr){6-6} \cmidrule(lr){7-7} \cmidrule(lr){8-9} \cmidrule(lr){10-11} \cmidrule(lr){12-13} \cmidrule(lr){14-14} \cmidrule(lr){15-15} \cmidrule(lr){16-16} \cmidrule(lr){17-17} \cmidrule(lr){18-18} \cmidrule(lr){19-19} \cmidrule(lr){20-20} \cmidrule(lr){21-22}
     & $P$ & 3 & 3 & 1 & 1 & 2 & 1 & 3 & 1 & 3 & 1 & 3 & 1 & 3 & 3 & 6 & 6 & 6 & 32 & 1 & 32 \\
    \midrule
    \multicolumn{20}{l}{\textbf{Regression-based models}} \\
    \multicolumn{2}{l}{Long-LRM~\cite{longlrm}} & - & - & - & - & - & - & - & - & - & - & - & - & - & - & - & - & - & \besttwo{0.262} & - & \besttwo{0.375} \\
     \multicolumn{2}{l}{MVSplat~\cite{chen2025mvsplat}} & \bestthree{\underline{0.411}} & \bestfour{\underline{0.387}} & \bestfour{\underline{0.224}} & \bestthree{\underline{0.237}} & \besttwo{0.128} & \besttwo{\underline{0.254}} & \besttwo{\underline{0.142}} & \bestthree{\underline{0.542}} & \bestfour{\underline{0.497}} & \bestthree{\underline{0.386}} & \bestfour{\underline{0.310}} & \underline{0.634} & \bestfour{\underline{0.614}}& \bestfour{\underline{0.504}} & \bestfour{\underline{0.643}} & \bestfour{\underline{0.556}} & \bestthree{\underline{0.527}} & \bestfour{\underline{0.425}} & \underline{0.519} & \bestfour{\underline{0.568}} \\
     \multicolumn{2}{l}{DepthSplat~\cite{depthsplat}} & \bestthree{\underline{0.404}} & \besttwo{\underline{0.372}} & \besttwo{\underline{0.217}} & \bestfour{\underline{0.245}} & \bestthree{0.119} & \best{\underline{0.236}} & \bestfour{\underline{0.177}}   & \besttwo{\underline{0.530}} & \bestthree{\underline{0.465}} & \besttwo{\underline{0.369}} & \bestthree{\underline{0.304}} & \bestfour{\underline{0.618}} & \bestthree{\underline{0.603}}& \bestthree{\underline{0.499}} & \besttwo{\underline{0.530}} & \bestthree{\underline{0.534}} & \besttwo{\underline{0.481}} & \bestthree{\underline{0.404}} & \bestthree{\underline{0.462}} & \bestthree{\underline{0.528}} \\
     \multicolumn{2}{l}{LVSM~\cite{jin2024lvsmlargeviewsynthesis}} & - & - & - & - & \besttwo{0.098} & - & - & - & - & -& - & -& - & -& - & -& - & - & - & - \\
    \midrule
    \multicolumn{20}{l}{\textbf{Diffusion-based models}} \\
     \multicolumn{2}{l}{MotionCtrl~\cite{chen2023motion}} & - & - & 0.500 & 0.386 & -  & - & - & - & - & - & - & \bestthree{0.443} & - & - & - & - & - & - & \bestfour{0.473} & - \\
     \multicolumn{2}{l}{4DiM~\cite{4dim}} & - & - & 0.302 & - & - & - & - & - & - & - & -& - & - & -& - & - & - & - & - & - \\
     \multicolumn{2}{l}{ViewCrafter~\cite{yu2024viewcrafter}} & \bestfour{\underline{0.427}} & \bestthree{\underline{0.379}} & \bestthree{\underline{0.220}} & \best{0.178} & \underline{0.203} & \bestthree{\underline{0.287}} & \bestthree{\underline{0.164}} & \bestfour{\underline{0.620}} & \besttwo{\underline{0.435}} & \bestfour{\underline{0.485}} & \besttwo{\underline{0.272}} & \besttwo{0.324} & \besttwo{\underline{0.513}} & \besttwo{\underline{0.324}} & \bestthree{\underline{0.639}} & \besttwo{\underline{0.464}} & \bestfour{\underline{0.558}} & - & \best{0.283} & - \\
     \multicolumn{2}{l}{\ours} & \best{0.049} & \best{0.041} & \best{0.194} & \besttwo{0.231} & \best{0.061} & \bestfour{0.308} & \best{0.073} & \best{0.389}  & \best{0.181} &  \best{0.316} & \best{0.158} & \best{0.318} & \best{0.278} & \best{0.215} & \best{0.237} & \best{0.319} & \best{0.232} & \best{0.155} & \besttwo{0.354} & \best{0.236} \\
    \bottomrule
    \end{tabular}
}
\vspace{0.2cm} %
\subfloat[\textbf{SSIM$\uparrow$}]{
    \begin{tabular}{lccccccccccccccccccccc}
    \toprule
    \multirow{3}{*}{Method} & dataset & OO3D & GSO & \multicolumn{5}{c}{RE10K} & \multicolumn{2}{c}{LLFF} & \multicolumn{2}{c}{DTU} & \multicolumn{2}{c}{CO3D} & \multicolumn{2}{c}{WRGBD} & Mip360 & \multicolumn{2}{c}{DL3DV} & \multicolumn{2}{c}{T\&T} \\
    \cmidrule(lr){2-2} \cmidrule(lr){3-3} \cmidrule(lr){4-4} \cmidrule(lr){5-9} \cmidrule(lr){10-11} \cmidrule(lr){12-13} \cmidrule(lr){14-15} \cmidrule(lr){16-17} \cmidrule(lr){18-18} \cmidrule(lr){19-20} \cmidrule(lr){21-22}
     & split & O & O & D~\cite{4dim} & V~\cite{yu2024viewcrafter} & P~\cite{charatan23pixelsplat} & \multicolumn{2}{c}{R~\cite{reconfusion}}   & \multicolumn{2}{c}{R~\cite{reconfusion}} & \multicolumn{2}{c}{R~\cite{reconfusion}} & V~\cite{yu2024viewcrafter} & R~\cite{reconfusion} & O$_\mathrm{e}$ & O$_\mathrm{h}$ & R~\cite{reconfusion} & O & L~\cite{longlrm} & V~\cite{yu2024viewcrafter} & L~\cite{longlrm} \\
     \cmidrule(lr){2-2} \cmidrule(lr){3-3} \cmidrule(lr){4-4} \cmidrule(lr){5-5} \cmidrule(lr){6-6} \cmidrule(lr){7-7} \cmidrule(lr){8-9} \cmidrule(lr){10-11} \cmidrule(lr){12-13} \cmidrule(lr){14-14} \cmidrule(lr){15-15} \cmidrule(lr){16-16} \cmidrule(lr){17-17} \cmidrule(lr){18-18} \cmidrule(lr){19-19} \cmidrule(lr){20-20} \cmidrule(lr){21-22}
     & $P$ & 3 & 3 & 1 & 1 & 2 & 1 & 3 & 1 & 3 & 1 & 3 & 1 & 3 & 3 & 6 & 6 & 6 & 32 & 1 & 32 \\
    \midrule
    \multicolumn{20}{l}{\textbf{Regression-based models}} \\
    \multicolumn{2}{l}{Long-LRM~\cite{longlrm}} & - & - & - & - & - & - & - & - & - & - & - & - & - & - & - & - & - & \best{0.775} & - & \best{0.590} \\
     \multicolumn{2}{l}{MVSplat~\cite{chen2025mvsplat}} & \bestthree{\underline{0.554}} & \bestfour{\underline{0.621}} & \bestthree{\underline{0.788}} & \bestthree{\underline{0.769}} & \bestthree{0.869} & \besttwo{\underline{0.812}} & \besttwo{\underline{0.857}} & \bestthree{\underline{0.283}} & \bestfour{\underline{0.358}} & \bestthree{\underline{0.576}} & \bestfour{\underline{0.624}} & \underline{0.403} & \bestthree{\underline{0.370}}& \bestfour{\underline{0.405}} & \bestfour{\underline{0.368}} & \bestfour{\underline{0.312}} & \bestthree{\underline{0.487}} & \bestfour{\underline{0.512}} & \underline{0.394} & \bestfour{\underline{0.314}} \\
     \multicolumn{2}{l}{DepthSplat~\cite{depthsplat}} & \besttwo{\underline{0.636}} & \besttwo{\underline{0.689}} & \best{\underline{0.801}} & \besttwo{\underline{0.745}} & \besttwo{0.887} & \best{\underline{0.820}} & \bestfour{\underline{0.824}}   & \besttwo{\underline{0.299}} & \bestthree{\underline{0.396}} & \besttwo{\underline{0.601}} & \bestthree{\underline{0.638}} & \bestfour{\underline{0.429}} & \besttwo{\underline{0.402}}& \bestthree{\underline{0.436}} & \besttwo{\underline{0.417}} & \bestthree{\underline{0.324}} & \besttwo{\underline{0.513}} & \bestthree{\underline{0.564}} & \bestthree{\underline{0.413}} & \bestthree{\underline{0.359}} \\
     \multicolumn{2}{l}{LVSM~\cite{jin2024lvsmlargeviewsynthesis}} & - & - & - & - & \best{0.906} & - & - & - & - & -& - & -& - & -& - & -& - & - & - & - \\
    \midrule
    \multicolumn{20}{l}{\textbf{Diffusion-based models}} \\
     \multicolumn{2}{l}{MotionCtrl~\cite{chen2023motion}} & - & - & 0.267 & 0.587 & -  & - & - & - & - & - & - & \bestthree{0.502} & - & - & - & - & - & - & \bestfour{0.384} & - \\
     \multicolumn{2}{l}{4DiM~\cite{4dim}} & - & - & 0.463 & - & - & - & - & - & - & - & -& - & - & -& - & - & - & - & - & - \\
     \multicolumn{2}{l}{ViewCrafter~\cite{yu2024viewcrafter}} & \bestfour{\underline{0.538}} & \bestthree{\underline{0.647}} & \besttwo{\underline{0.792}} & \best{0.798} & \underline{0.710} & \bestthree{\underline{0.806}} & \bestthree{\underline{0.830}} & \bestfour{\underline{0.146}} & \besttwo{\underline{0.454}} & \bestfour{\underline{0.542}} & \besttwo{\underline{0.671}} & \best{0.641} & \bestfour{\underline{0.483}} & \besttwo{\underline{0.465}} & \bestthree{\underline{0.376}} & \besttwo{\underline{0.354}} & \bestfour{\underline{0.469}} & - & \best{0.563} & - \\
     \multicolumn{2}{l}{\ours} & \best{0.935} & \best{0.942} & \bestfour{0.615} & \bestfour{0.693} & \bestfour{0.847} & \bestfour{0.700} & \best{0.892} & \best{0.384}  & \best{0.602} &  \best{0.652} & \best{0.750} & \besttwo{0.585} & \best{0.647} & \best{0.670} & \best{0.646} & \best{0.395} & \best{0.546} & \besttwo{0.661} & \besttwo{0.437} & \besttwo{0.505} \\
    \bottomrule
    \end{tabular}
}
\caption{
\textbf{LPIPS$\downarrow$ (top) and SSIM$\uparrow$ (bottom) on small-viewpoint set NVS.}
For all results with $P=1$, we sweep the unit length for camera normalization due to the model's scale ambiguity.
O$_e$ and O$_h$ denote the easy and hard split of our split.
\underline{Underlined} numbers are run by us using the officially released code.
}
\label{tab:setnvs_small_app}
\end{table*}

\begin{table*}[ht!]
    \centering
    \begin{minipage}{0.63\textwidth}
        \centering
        \tablestyle{0.5pt}{1.0}
        \subfloat[\textbf{LPIPS$\downarrow$}]{
        \begin{tabular}{lccccccccccccccc}
            \toprule
            \multirow{3}{*}{Method} & dataset & OO3D & GSO & CO3D & \multicolumn{2}{c}{WRGBD} & \multicolumn{2}{c}{Mip360} & \multicolumn{2}{c}{DL3DV} & \multicolumn{4}{c}{T\&T} \\
            \cmidrule(lr){2-2} \cmidrule(lr){3-3} \cmidrule(lr){4-4} \cmidrule(lr){5-5} \cmidrule(lr){6-7} \cmidrule(lr){8-9} \cmidrule(lr){10-11} \cmidrule(lr){12-15}
             & split & S~\cite{voleti2024sv3d} & S~\cite{voleti2024sv3d} & R~\cite{reconfusion} & \multicolumn{2}{c}{O$_\mathrm{h}$} & \multicolumn{2}{c}{R~\cite{reconfusion}} & \multicolumn{2}{c}{O} & \multicolumn{4}{c}{O} \\
             \cmidrule(lr){2-2} \cmidrule(lr){3-3} \cmidrule(lr){4-4} \cmidrule(lr){5-5} \cmidrule(lr){6-7} \cmidrule(lr){8-9} \cmidrule(lr){10-11} \cmidrule(lr){12-15}
             & $P$ & 1 & 1 & 1 & 1 & 3 & 1 & 3 & 1 & 3 & 1 & 3 & 6 & 9 \\
            \midrule
            \multicolumn{2}{l}{SV3D~\cite{voleti2024sv3d}} & \best{0.158} & \besttwo{0.140} & - & - & - & - & - & - & - & - & - & - & -\\
             \multicolumn{2}{l}{DepthSplat~\cite{depthsplat}} & \bestthree{\underline{0.610}} & \bestthree{\underline{0.543}} & \besttwo{\underline{0.756}} & \besttwo{\underline{0.732}} & \besttwo{\underline{0.588}} & \besttwo{\underline{0.691}} & \bestthree{\underline{0.491}} & \besttwo{\underline{0.580}} & \besttwo{\underline{0.405}} & \bestthree{\underline{0.774}} & \bestthree{\underline{0.706}} & \bestthree{\underline{0.611}} & \besttwo{\underline{0.487}} \\
             \multicolumn{2}{l}{CAT3D~\cite{cat3d}} & - & - & - & - & - & - & \besttwo{0.488} & - & - & - & - & - & -\\
             \multicolumn{2}{l}{ViewCrafter~\cite{yu2024viewcrafter}} & \bestfour{\underline{0.634}} & \bestfour{\underline{0.559}} & \bestthree{\underline{0.789}} & \bestthree{\underline{0.775}} & \bestthree{\underline{0.603}} & \bestthree{\underline{0.723}} & \bestfour{\underline{0.540}} & \bestthree{\underline{0.616}} & \bestthree{\underline{0.576}} & \besttwo{\underline{0.755}} & \besttwo{\underline{0.671}} & \besttwo{\underline{0.604}} & \bestthree{\underline{0.546}} \\
             \multicolumn{2}{l}{\ours} & \besttwo{0.160} & \best{0.137} & \best{0.445} & \best{0.423} & \best{0.289} & \best{0.573} & \best{0.364} & \best{0.484} & \best{0.316} & \best{0.571} & \best{0.463} & \best{0.387} & \best{0.328} \\
            \bottomrule
        \end{tabular}}
        \vspace{0.2cm} %
        \subfloat[\textbf{SSIM$\uparrow$}]{
        \begin{tabular}{lccccccccccccccc}
            \toprule
            \multirow{3}{*}{Method} & dataset & OO3D & GSO & CO3D & \multicolumn{2}{c}{WRGBD} & \multicolumn{2}{c}{Mip360} & \multicolumn{2}{c}{DL3DV} & \multicolumn{4}{c}{T\&T} \\
            \cmidrule(lr){2-2} \cmidrule(lr){3-3} \cmidrule(lr){4-4} \cmidrule(lr){5-5} \cmidrule(lr){6-7} \cmidrule(lr){8-9} \cmidrule(lr){10-11} \cmidrule(lr){12-15}
             & split & S~\cite{voleti2024sv3d} & S~\cite{voleti2024sv3d} & R~\cite{reconfusion} & \multicolumn{2}{c}{O$_\mathrm{h}$} & \multicolumn{2}{c}{R~\cite{reconfusion}} & \multicolumn{2}{c}{O} & \multicolumn{4}{c}{O} \\
             \cmidrule(lr){2-2} \cmidrule(lr){3-3} \cmidrule(lr){4-4} \cmidrule(lr){5-5} \cmidrule(lr){6-7} \cmidrule(lr){8-9} \cmidrule(lr){10-11} \cmidrule(lr){12-15}
             & $P$ & 1 & 1 & 1 & 1 & 3 & 1 & 3 & 1 & 3 & 1 & 3 & 6 & 9 \\
            \midrule
            \multicolumn{2}{l}{SV3D~\cite{voleti2024sv3d}} & \besttwo{0.850} & \best{0.880} & - & - & - & - & - & - & - & - & - & - & -\\
             \multicolumn{2}{l}{DepthSplat~\cite{depthsplat}} & \bestthree{\underline{0.549}} & \bestthree{\underline{0.612}} & \besttwo{\underline{0.385}} & \besttwo{\underline{0.234}} & \besttwo{\underline{0.335}} & \besttwo{\underline{0.206}} & \bestthree{\underline{0.291}} & \besttwo{\underline{0.349}} & \besttwo{\underline{0.452}} & \bestthree{\underline{0.304}} & \bestthree{\underline{0.315}} & \bestthree{\underline{0.326}} & \besttwo{\underline{0.367}} \\
             \multicolumn{2}{l}{CAT3D~\cite{cat3d}} & - & - & - & - & - & - & \besttwo{0.294} & - & - & - & - & - & -\\
             \multicolumn{2}{l}{ViewCrafter~\cite{yu2024viewcrafter}} & \bestfour{\underline{0.463}} & \bestfour{\underline{0.575}} & \bestthree{\underline{0.277}} & \bestthree{\underline{0.225}} & \bestthree{\underline{0.321}} & \bestthree{\underline{0.199}} & \bestfour{\underline{0.264}} & \bestthree{\underline{0.323}} & \bestthree{\underline{0.400}} & \besttwo{\underline{0.312}} & \besttwo{\underline{0.328}} & \besttwo{\underline{0.337}} & \bestthree{\underline{0.343}} \\
             \multicolumn{2}{l}{\ours} & \best{0.857} & \besttwo{0.873} & \best{0.536} & \best{0.505} & \best{0.603} & \best{0.282} & \best{0.377} & \best{0.360} & \best{0.480} & \best{0.342} & \best{0.385} & \best{0.427} & \best{0.452} \\
            \bottomrule
        \end{tabular}}
        \caption{
        \textbf{LPIPS$\downarrow$ (top) and SSIM$\uparrow$ (bottom) on large-viewpoint set NVS.} 
        For all results with $P=1$, we sweep the unit length for camera normalization due to the model's scale ambiguity. 
        \underline{Underlined} numbers are run by us using the officially released code.
        }
        \label{tab:setnvs_large_app}
    \end{minipage}
    \hfill
    \begin{minipage}{0.36\textwidth}
        \centering
        \tablestyle{0.8pt}{1.0}
        \subfloat[\textbf{LPIPS$\downarrow$}]{
        \begin{tabular}{lccccc}
        \toprule
        \multirow{2}{*}{Method} & \multicolumn{4}{c}{small-viewpoint} & \makecell{large-\\viewpoint} \\
        \cmidrule(lr){2-5} \cmidrule(lr){6-6}
        & RE10K & LLFF & DTU & CO3D & Mip360 \\
        \midrule
        ZipNeRF~\cite{barron2023zip} & \bestfour{0.332}	& \bestfour{0.373} & 0.383 & 0.652 & 0.705 \\
        ZeroNVS~\cite{sargent2023zeronvs} & 0.422	& 0.512 & \bestfour{0.223} & \bestfour{0.566} & \bestfour{0.680} \\
        ReconFusion~\cite{reconfusion} & \bestthree{0.144}	& \bestthree{0.203} & \bestthree{0.124} & \bestthree{0.398} & \bestthree{0.585} \\
        CAT3D~\cite{cat3d} & \besttwo{0.132} & \besttwo{0.181} & \besttwo{0.121} & \besttwo{0.351} &	\besttwo{0.515} \\
        \ours & \best{0.078} & \best{0.164} & \best{0.107} & \best{0.256} & \best{0.435} \\
        \bottomrule
        \end{tabular}}
        \vspace{0.2cm} %
        \subfloat[\textbf{SSIM$\uparrow$}]{
        \begin{tabular}{lccccc}
        \toprule
        \multirow{2}{*}{Method} & \multicolumn{4}{c}{small-viewpoint} & \makecell{large-\\viewpoint} \\
        \cmidrule(lr){2-5} \cmidrule(lr){6-6}
        & RE10K & LLFF & DTU & CO3D & Mip360 \\
        \midrule
        ZipNeRF~\cite{barron2023zip} & \bestfour{0.774}	& \bestfour{0.574} & 0.601 & 0.496 & 0.271 \\
        ZeroNVS~\cite{sargent2023zeronvs} & 0.675	& 0.359 & \bestfour{0.716} & \bestfour{0.581} & \bestfour{0.316} \\
        ReconFusion~\cite{reconfusion} & \bestthree{0.910}	& \bestthree{0.724} & \best{0.875} & \bestthree{0.662} & \bestthree{0.358} \\
        CAT3D~\cite{cat3d} & \besttwo{0.917} & \besttwo{0.731} & \bestthree{0.844} & \besttwo{0.666} &	\besttwo{0.377} \\
        \ours & \best{0.961} & \best{0.735} & \besttwo{0.867} & \best{0.702} & \best{0.454} \\
        \bottomrule
        \end{tabular}}
        \caption{
        \textbf{LPIPS$\downarrow$ (top) and SSIM$\uparrow$ (bottom) on 3DGS renderings for set NVS.} Results are reported on the ReconFusion~\cite{reconfusion} split with $P=3$.
        }
        \label{tab:3dgs_app}
    \end{minipage}
\end{table*}

\begin{table}
\tablestyle{1.6pt}{1.0}
\centering
\subfloat[\textbf{LPIPS$\downarrow$}]{
\begin{tabular}{lccccccccc}
    \toprule
    \multirow{4}{*}{Method} & \multirow{2}{*}{split} & \multicolumn{5}{c}{small-viewpoint} & \multicolumn{3}{c}{large-viewpoint} \\
    \cmidrule(lr){3-7} \cmidrule(lr){8-10}
    & & \multicolumn{3}{c}{V~\cite{yu2024viewcrafter}} & \multicolumn{5}{c}{O} \\
    \cmidrule(lr){2-2} \cmidrule(lr){3-5} \cmidrule(lr){6-10}
    & dataset & RE & CO3D & T\&T & RE & DTU & WR & DL & T\&T \\
    \midrule
     \multicolumn{2}{l}{MotionCtrl~\cite{chen2023motion}} & 0.386 & \bestfour{0.443} & 0.473 & - & - & - & - & - \\
     \multicolumn{2}{l}{DepthSplat~\cite{depthsplat}} & \besttwo{\underline{0.224}} & \underline{0.532} & \bestfour{\underline{0.415}}  & \bestfour{\underline{0.134}} & \bestfour{\underline{0.253}} & \bestfour{\underline{0.452}} & \bestfour{\underline{0.572}} & \bestfour{\underline{0.685}} \\
     \multicolumn{2}{l}{ViewCrafter~\cite{yu2024viewcrafter}} & \best{0.178} & \best{0.283} & \best{0.324} & \bestthree{\underline{0.120}} & \bestthree{\underline{0.187}} & \bestthree{\underline{0.346}} & \bestthree{\underline{0.566}} & \bestthree{\underline{0.674}} \\
     \midrule 
     \multicolumn{2}{l}{\ours} & \bestfour{0.231} & \bestthree{0.318} & \besttwo{0.353} & \besttwo{0.079} & \besttwo{0.159} & \besttwo{0.284} & \besttwo{0.329} & \besttwo{0.514} \\
     \multicolumn{2}{l}{\ours (+ temp.)} & \bestthree{0.228} & \besttwo{0.312} & \bestthree{0.356} & \best{0.078} & \best{0.156} & \best{0.280} & \best{0.328} & \best{0.510} \\
    \bottomrule
\end{tabular}}
\vspace{0.2cm}
\subfloat[\textbf{SSIM$\uparrow$}]{
\begin{tabular}{lccccccccc}
    \toprule
    \multirow{4}{*}{Method} & \multirow{2}{*}{split} & \multicolumn{5}{c}{small-viewpoint} & \multicolumn{3}{c}{large-viewpoint} \\
    \cmidrule(lr){3-7} \cmidrule(lr){8-10}
    & & \multicolumn{3}{c}{V~\cite{yu2024viewcrafter}} & \multicolumn{5}{c}{O} \\
    \cmidrule(lr){2-2} \cmidrule(lr){3-5} \cmidrule(lr){6-10}
    & dataset & RE & CO3D & T\&T & RE & DTU & WR & DL & T\&T \\
    \midrule
     \multicolumn{2}{l}{MotionCtrl~\cite{chen2023motion}} & 0.587 & \bestfour{0.502} & 0.384 & - & - & - & - & - \\
     \multicolumn{2}{l}{DepthSplat~\cite{depthsplat}} & \besttwo{\underline{0.723}} & \underline{0.486} & \bestfour{\underline{0.408}}  & \bestfour{\underline{0.844}} & \bestfour{\underline{0.723}} & \bestfour{\underline{0.447}} & \bestfour{\underline{0.539}} & \bestfour{\underline{0.497}} \\
     \multicolumn{2}{l}{ViewCrafter~\cite{yu2024viewcrafter}} & \best{0.798} & \best{0.641} & \best{0.563} & \bestthree{\underline{0.868}} & \bestthree{\underline{0.739}} & \bestthree{\underline{0.464}} & \bestthree{\underline{0.523}} & \bestthree{\underline{0.456}} \\
     \midrule 
     \multicolumn{2}{l}{\ours} & \bestfour{0.693} & \bestthree{0.585} & \besttwo{0.437} & \besttwo{0.890} & \besttwo{0.756} & \besttwo{0.613} & \besttwo{0.475} & \besttwo{0.363} \\
     \multicolumn{2}{l}{\ours (+ temp.)} & \bestthree{0.695} & \besttwo{0.590} & \bestthree{0.436} & \best{0.891} & \best{0.760} & \best{0.616} & \best{0.476} & \best{0.369} \\
    \bottomrule
\end{tabular}}
\label{tab:trajnvs_app}
\caption{
\textbf{LPIPS$\downarrow$ (top) and SSIM$\uparrow$ (bottom) on trajectory NVS.} 
For the V~\cite{yu2024viewcrafter} split, $P=1$ with unit length swept; for the O split, $P=3$. 
RE, WR, and DL denote RE10K, WRGBD, and DL3DV, respectively.
\underline{Underlined} numbers are run by us using the officially released code.
}
\end{table}

We collect 10 commonly used datasets to benchmark NVS, encompassing a diverse range of scene distributions and complexities, shown in~\cref{tab:benchmark_dataset}.

\paragraph{Small-viewpoint \textit{versus} large-viewpoint NVS.}
In~\cref{sec:benchmark}, we split NVS tasks into two categories: small-viewpoint and large-viewpoint NVS based on the disparity between $\rmI^\text{inp}$ and $\rmI^\text{tgt}$.
Formally, for each target view, we consider the minimal distance between the CLIP~\cite{radford2021clip} feature of that view and those of all input views. Averaging across all target views yields the CLIP distance, $\mathcal{D}_\mathrm{CLIP}(\rmI)$.
Splits with $\mathcal{D}_\mathrm{CLIP}(\rmI)<=0.11$ are grouped as small-view NVS, while those with $\mathcal{D}_\mathrm{CLIP}(\rmI)>0.11$ are grouped as large-view NVS.
We concrete in~\cref{tab:benchmark_dataset} a detailed task setup including the choice of datasets and splits (depending on which scenes from each dataset and which views from each scene are used). 

\paragraph{Choice of scenes.}
We follow the choices of scenes for splits adopted from previous works. For our split, we use all scenes from the dataset without specification. 

For the Tanks and Temples dataset, the 2 chosen scenes in our (O) split are \textsc{Train} and \textsc{Truck}. 
For the DL3DV-140 dataset, the 10 test scenes we choose in O split are:
\begin{itemize}[noitemsep,topsep=0pt,leftmargin=*]
    \item \textsc{165f5af8bfe32f70595a1c9393a6e442acf7af 019998275144f605b89a306557}
    \item \textsc{341b4ff3dfd3d377d7167bd81f443bedafbff0 03bf04881b99760fc0aeb69510}
    \item \textsc{3bb3bb4d3e871d79eb71946cbab1e3afc7a8e3 3a661153033f32deb3e23d2e52}
    \item \textsc{3bb894d1933f3081134ad2d40e54de5f0636bd 8b502b0a8561873bb63b0dce85}
    \item \textsc{9e9a89ae6fed06d6e2f4749b4b0059f35ca97f 848cedc4a14345999e746f7884}
    \item \textsc{cd9c981eeb4a9091547af19181b382698e9d9e ee0a838c7c9783a8a268af6aee}
    \item \textsc{d4fbeba0168af8fddb2fc695881787aedcd62f 477c7dcec9ebca7b8594bbd95b}
    \item \textsc{e78f8cebd2bd93d960bfaeac18fac0bb2524f1 5c44288903cd20b73e599e8a81}
    \item \textsc{ed16328235c610f15405ff08711eaf15d88a05 03884f3a9ccb5a0ee69cb4acb5}
    \item \textsc{f71ac346cd0fc4652a89afb37044887ec3907d 37d01d1ceb0ad28e1a780d8e03}. 
\end{itemize}
For the WildRGBD dataset, the 20 test scenes we choose in O split are: 
For the WildRGBD dataset, the 20 test scenes we select for the O split are:
\begin{itemize}[noitemsep,topsep=0pt,leftmargin=*]
    \item \textsc{ball/scene\_563}
    \item \textsc{apple/scene\_234}
    \item \textsc{microwave/scene\_143}
    \item \textsc{scissor/scene\_489}
    \item \textsc{bucket/scene\_294}
    \item \textsc{keyboard/scene\_092}
    \item \textsc{shoe/scene\_868}
    \item \textsc{kettle/scene\_399}
    \item \textsc{clock/scene\_524}
    \item \textsc{hat/scene\_039}
    \item \textsc{backpack/scene\_264}
    \item \textsc{scissor/scene\_958}
    \item \textsc{truck/scene\_232}
    \item \textsc{handbag/scene\_575}
    \item \textsc{pineapple/scene\_182}
    \item \textsc{train/scene\_033}
    \item \textsc{remote\_control/scene\_453}
    \item \textsc{bowl/scene\_673}
    \item \textsc{TV/scene\_062}
\end{itemize}
Full test scenes are chosen for the remaining datasets.

\paragraph{Choice of input and target views.} 
We follow the same setup for splits adopted from previous works, by using the same set of input and target views. 
For split defined ourselves, we detail the choice of views as below.
For the WildRGB-D~\cite{wildrgbd} dataset, which consists of scenes captured while orbiting around an object, we define two splits with different difficulty levels. O$_e$ represents the easy set, where each scene is trimmed to one-third of the original sequence (\ie, approximately 120 degrees of rotation). In contrast, O$_h$ corresponds to the hard set, using the full original sequence (\ie, approximately 360 degrees of rotation).
We first uniformly subsample 21 frames from the scene, and randomly choose $P$ frames as input views with the remaining frames as target views.
For each scene from DL3DV-140~\cite{dl3dv10k} and Tanks and Temples~\cite{knapitsch2017tanks} datasets, we selected target frames by using every 8$^\mathrm{th}$ frame of the original sequence. For the remaining frames, we applied $K$-means clustering ($K=32$) on a 6-dimensional vector formed by concatenating the camera translation and the unit vector of the camera direction.

\section{Additional Experiments}
\label{app:addexperiments}

\subsection{Qualitative Results}
\label{app:qualitative}

We provide additional single-view conditioning sampling results with a diverse set of camera motions and effects on a variety of image prompts: a text-prompted object-centric scene (\cref{fig:1view_obj_prompt}), a text-prompted scene (\cref{fig:1view_scene_prompt}), a real-world object-centric scene (\cref{fig:1view_obj}), and a real-world scene (\cref{fig:1view_scene}). \ours demonstrates strong generalization, adapting robustly across a wide range of scenarios.

\subsection{Quantitative Results}
\label{app:quantitative}
We provide additional quantitative evaluation results of our model against baselines on set NVS and trajectory NVS, measured using LPIPS~\cite{zhang2018lpips} and SSIM~\cite{wang2004image}, in~\cref{tab:setnvs_large_app,tab:setnvs_small_app,tab:3dgs_app,tab:trajnvs_app}.

\subsection{Discussion}
\label{app:discussion}

\begin{figure}
    \includegraphics[width=\linewidth]{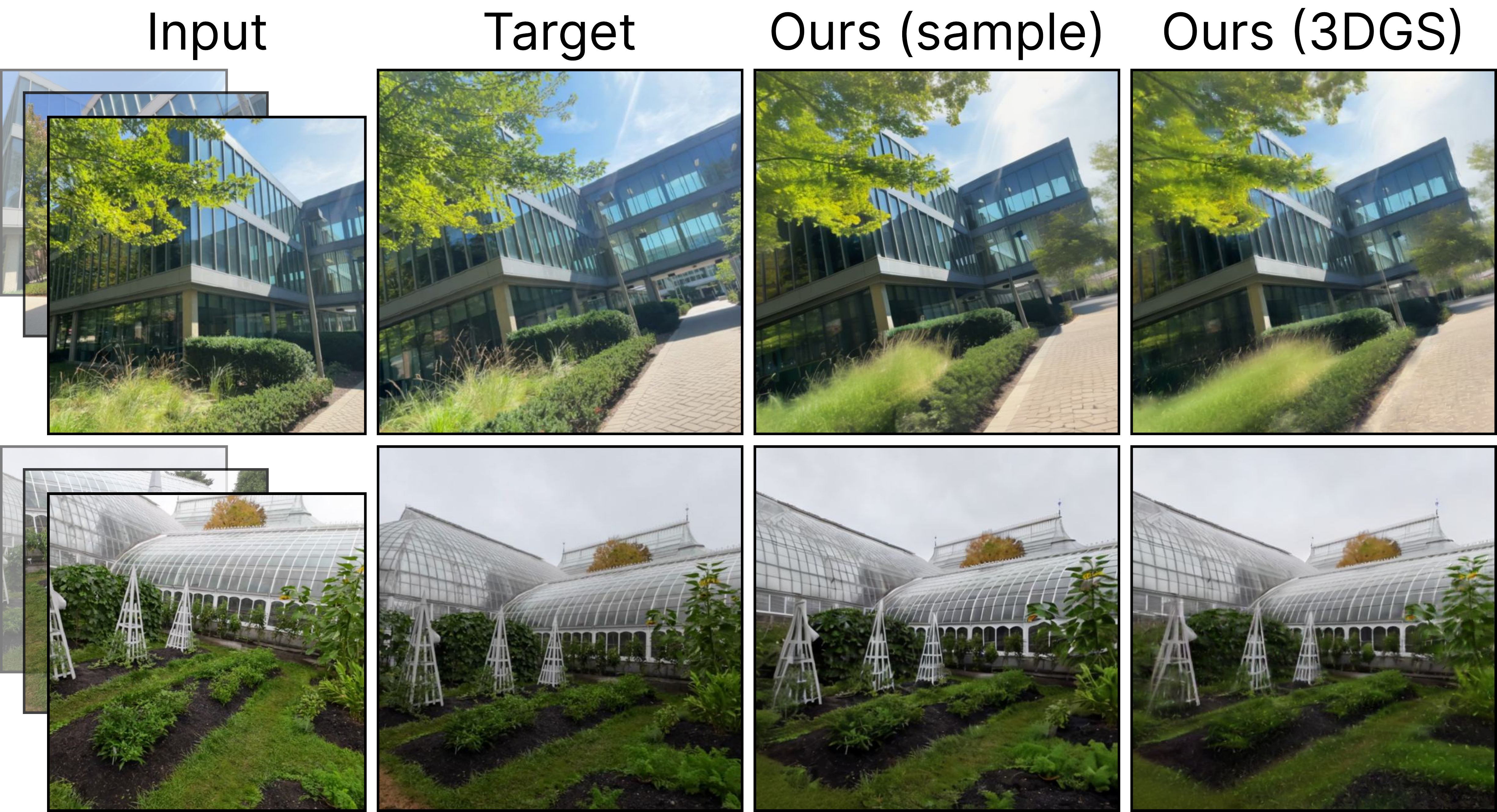}
    \captionof{figure}{
        \textbf{3DGS versus samples.}
        The model generates consistent renderings that closely resemble those from 3DGS~\cite{kerbl3Dgaussians}, with minimal perceptual differences.
    }
    \label{fig:ablation-3dgs-vs-sample}
\end{figure}

\paragraph{Samples \textit{versus} 3DGS.}
We compare our samples to their 3DGS distillation on the O split of DL3DV, shown in~\cref{fig:ablation-3dgs-vs-sample}.
First, we note that our samples contain plausible hallucinations when uncertainty is high (first row, building on the right).
Second, we note that our 3DGS renderings remain sharp and are close to the samples.
These results suggest that our samples are 3D consistent enough.

\begin{figure}
    \includegraphics[width=\linewidth]{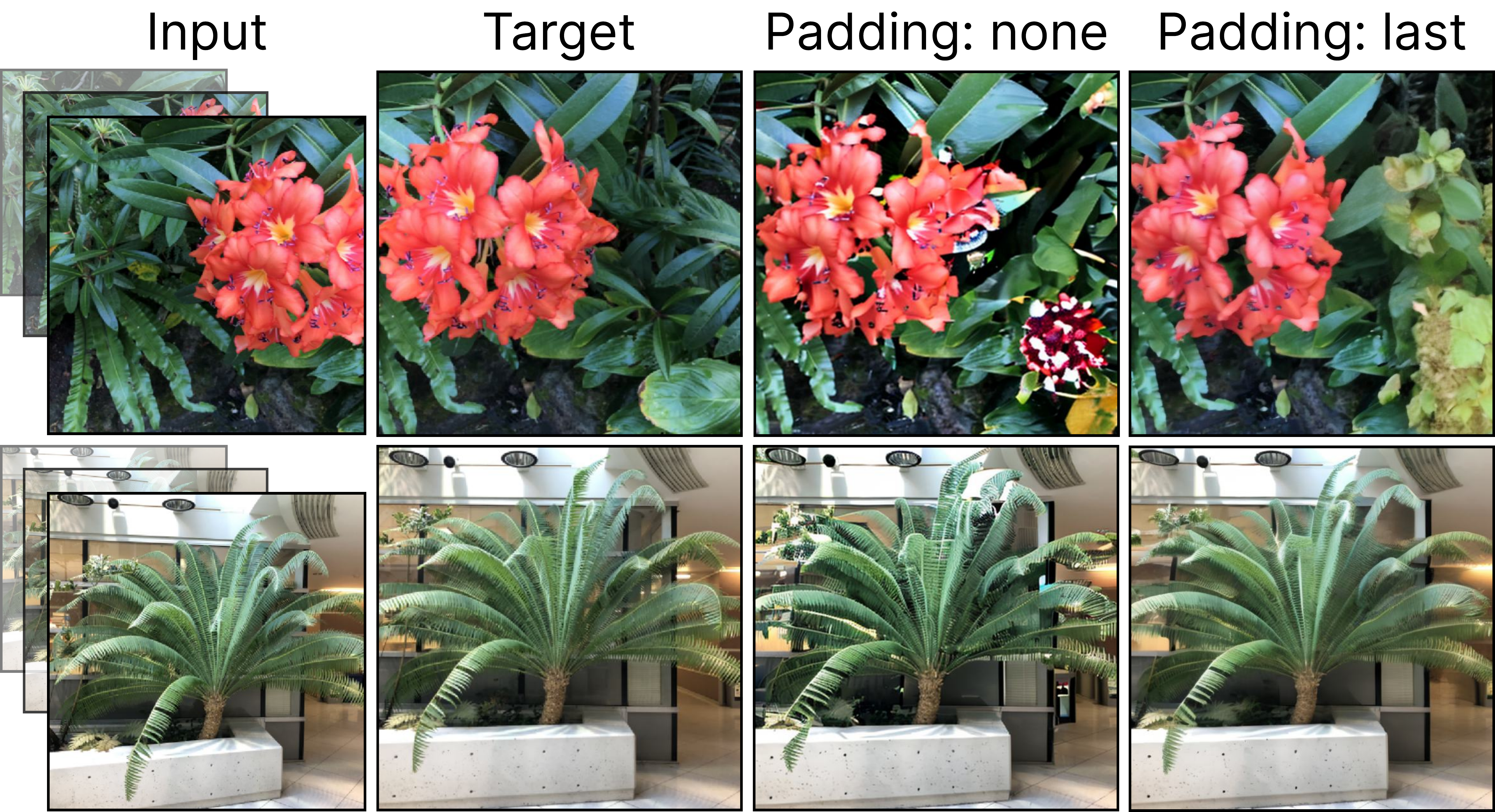}
    \captionof{figure}{
        \textbf{Padding.}
        Padding the last elements within one forward reduces artifacts compared to changing $T$.
    }
    \label{fig:ablation-padding}
\end{figure}

\paragraph{Padding $T$ when $P+Q<T$.}
We analyze the effect of different padding strategies when $P+Q<T$ in~\cref{fig:ablation-padding}. We observe that zero-shot generalization of $T$ to $P+Q$ without padding leads to abnormal color overflows.  
This is in stark contrast to the excessive blurriness observed when generalizing $T$ when $P+Q>>T$ in sparse-view regime (\cref{sec:discussions}).  
Hypothetically, sampling with a $T$ unseen during training induces a distribution shift in the attention scores~\cite{velivckovic2024softmax}. Specifically, a smaller $T$ sharpens the attention distribution, whereas a larger $T$ disperses it. This shift may explain the contrasting behavior observed when using the model for sampling. 
Training the model with a dynamically varying $T$ during training could mitigate this issue by exposing the model to a broader range of attention score distributions, improving generalization across different $T$.

\paragraph{Artifacts on long-trajectory NVS.}
We observe that the results tend to become increasingly saturated, particularly when the target views are far from the input views and share no content overlap, such as in open-ended exploration and navigation.
The concurrent work~\cite{song2025historyguidedvideodiffusion} explores the concept of Diffusion Forcing~\cite{chen2024diffusionforcingnexttokenprediction} for long video rollouts, achieving high-generation quality.
Applying diverse noise to the input views during training can be beneficial, as it enables the refinement of high-level details in all anchor views within the memory bank during sampling, thereby mitigating the accumulation of saturation.
We leave this for future work.

\begin{figure*}
    \includegraphics[width=0.95\textwidth]{fig/assets/result_1view_obj_prompt.pdf}
    \captionof{figure}{
        \textbf{Diverse camera motions and effects.} 
        Single-view conditioning with a text-prompted object-centric scene.
        The image is generated using SD 3.5~\cite{sd3} with the text prompt, ``\textit{A cute firefly dragon in its natural habitat.}''
    }
    \label{fig:1view_obj_prompt}
\end{figure*}

\begin{figure*}
    \includegraphics[width=0.95\textwidth]{fig/assets/result_1view_scene_prompt.pdf}
    \captionof{figure}{
        \textbf{Diverse camera motions and effects.} 
        Single-view conditioning with a text-prompted scene.
        The image is generated using SD 3.5~\cite{sd3} with the text prompt, ``\textit{Wide view of the interior of the famed Library of Alexandria, elegantly set behind a time-worn wreckage by a lake, hinting at the relentless passage of time. The surroundings are lit by the light of a late afternoon sun, gently cast, immersing the area in a sentimental luminescence. 
        }''
    }
    \label{fig:1view_scene_prompt}
\end{figure*}

\begin{figure*}
    \includegraphics[width=0.95\textwidth]{fig/assets/result_1view_obj.pdf}
    \captionof{figure}{
        \textbf{Diverse camera motions and effects.} 
        Single-view conditioning with a real-life object-centric scene.
    }
    \label{fig:1view_obj}
\end{figure*}

\begin{figure*}
    \includegraphics[width=0.95\textwidth]{fig/assets/result_1view_scene.pdf}
    \captionof{figure}{
        \textbf{Diverse camera motions and effects.} 
        Single-view conditioning with a real-life scene.
    }
    \label{fig:1view_scene}
\end{figure*}

\end{document}